\documentclass[11pt,letterpaper]{article}
\usepackage{amsfonts}
\usepackage{amsmath}
\usepackage{amsthm}
\usepackage{amssymb}
\usepackage{geometry}
\usepackage{indentfirst}
\usepackage{changepage}
\usepackage{graphicx}
\usepackage{epstopdf} 
\usepackage{color}
\usepackage{url}            % simple URL typesetting
\usepackage{mathtools}
\usepackage{algorithm}
\usepackage{algorithmic}
\usepackage{subfigure}
\usepackage{array}
\usepackage{floatflt}
\usepackage{multirow}
\usepackage{wrapfig}
\usepackage{microtype}  
\title{\textbf{Safe Element Screening for Submodular\\ Function Minimization}}
\author{Weizhong Zhang$^{1}$, Bin Hong$^{2}$, Lin Ma$^{1}$, Wei Liu$^{1}$, Tong Zhang$^{1}$\\
		$^1$Tencent AI Lab, Shenzhen, China\\
	$^2$State Key Lab of CAD$\&$CG, College of Computer Science, Zhejiang University
}

\def \c {\mathbf{c}}
\def \g {\mathbf{g}}

\def \s {\mathbf{s}}
\def \u {\mathbf{u}} 
\def \v {\mathbf{v}}
\def \w {\mathbf{w}}
\def \x {\mathbf{x}}
\def \y {\mathbf{y}}

\def \R {\mathbb{R}}
\def \calA {\mathcal{A}}
\def \calB {\mathcal{B}}

\def \calE {\mathcal{E}}

\def \calG {\mathcal{G}}

\def \calP {\mathcal{P}}

\def \calW {\mathcal{W}}

\def \hcalE {\hat{\mathcal{E}}}

\def \hcalG {\hat{\mathcal{G}}}

\newtheorem{theorem}{Theorem}
\newtheorem{lemma}{Lemma}
\newtheorem{definition}{Definition}
\newtheorem{remark}{Remark}

\renewcommand{\eqref}[1]{Eq.~(\ref{#1})}

\begin{document}
	\maketitle	
\begin{abstract} 
	Submodular functions are discrete analogs of convex functions, which have applications in various fields, including machine learning and computer vision. However, in large-scale applications, solving Submodular Function Minimization (SFM) problems remains challenging. In this paper, we make the first attempt to extend the emerging technique named screening in large-scale sparse learning to SFM for accelerating its optimization process. We first conduct a careful studying of the relationships between SFM and the corresponding convex proximal problems, as well as the accurate primal optimum estimation of the proximal problems. Relying on this study, we subsequently propose a novel safe screening method to quickly identify the elements guaranteed to be included (we refer to them as active) or excluded (inactive) in the final optimal solution of SFM during the optimization process. By removing the inactive elements and fixing the active ones, the problem size can be dramatically reduced, leading to great savings in the computational cost without sacrificing any accuracy. To the best of our knowledge, the proposed method is the first screening method in the fields of SFM and even combinatorial optimization, thus pointing out a new direction for accelerating SFM algorithms. Experiment results on both synthetic and real datasets demonstrate the significant speedups gained by our approach. 
\end{abstract} 

\section{Introduction} \label{sec:introduction}
Submodular Functions \cite{fujishige2005submodular} are a special class of set functions, which have rich structures and a lot of links with convex functions. They arise naturally in many domains, such as clustering \cite{Narasimhan:2007}, image segmentation \cite{kolmogorov2004energy,cevher2009sparse}, document summarization \cite{lin2011class}, etc. Most of these applications can be finally deduced to a Submodular Function Minimization (SFM) problem:
\begin{align}
\min_{A \subseteq V} F(A), \tag{SFM} \label{eqn:SFM}
\end{align}
where $F(A)$ is a submodular function defined on a set $V$. The problem of SFM has been extensively studied for several decades in the literatures \cite{edmonds1970submodular, lovasz1983submodular, mccormick2005submodular, wu2016constrained, ene2017decomposable}, in which many algorithms have been developed from the perspectives of combinatorial optimization and convex optimization. The most well-known conclusion is that SFM is solvable in strongly polynomial time \cite{iwata2001combinatorial}. Unfortunately, due to the high-degree polynomial dependence, the applications of submodular functions on the large scale problems remain challenging, such as image segmentation \cite{cevher2009sparse} and speech analysis \cite{lin2011optimal}, which both involve a huge number of variables.

Screening \cite{EVR:10} is an emerging technique, which has been proved to be effective in accelerating large-scale sparse model training.  It is motivated by the well-known feature of sparse models that a significant portion of the coefficients in the optimal solutions of them (resp. their dual problems) are zeros, that is, the corresponding features (resp. samples) are irrelevant with the final learned models. Screening methods aim to quickly identify these irrelevant features and/or samples and remove them from the datasets before or during the training process. Thus, the problem size can be reduced dramatically, leading to substantial savings in the computational cost. The framework of these methods is given in Algorithm \ref{alg:scrrening-framework}. Since screening methods are always independent of the training algorithms, they can be integrated with all the algorithms flexibly. In the recent few years, specific screening methods for most of the traditional sparse models have been developed, such as Lasso \cite{tibshirani2012strong, wang2013lasso, wang2015multi}, sparse logistic regression \cite{wang2014safe}, multi-task learning \cite{ndiaye2015gap}  and SVM \cite{ogawa2014safe,zhang2017scaling}. Empirical studies indicate that the speedups they achieved can be orders of magnitudes.
\begin{algorithm}[htb]
	\caption{Framework of screening in sparse learning}
	\begin{algorithmic}[1]
		\STATE Estimate the dual (resp. primal) optimum of the sparse model.
		\STATE Based on the estimation above, infer which components of the primal (resp. dual ) optimum are zeros from the KKT conditions.
		\STATE Remove the features (resp. samples) corresponding to the identified components.
		\STATE Train the model on the reduced dataset. 
	\end{algorithmic}\label{alg:scrrening-framework}
\end{algorithm}

The binary attribute (each element in $V$ must be either in or not in the optimal solution) of SFM  motivates us to introduce the key idea of screening into SFM to accelerate its optimization process. The most intuitive approach is to identify the elements that are guaranteed to be included or excluded in the minimizer $A^*$ of \ref{eqn:SFM} prior to or during actually solving it. Then, by fixing the identified active elements and removing the inactive ones, we just need to solve a small-scale problem. However, we note that existing screening methods are all developed for convex models and they cannot be applied to SFM directly. The reason is that they all heavily depend on KKT conditions (see Algorithm \ref{alg:scrrening-framework}), which do not exist in SFM problems.

In this paper, to improve the efficiency of SFM algorithms, we propose a novel \textbf{I}nactive and \textbf{A}ctive \textbf{E}lement \textbf{S}creening (IAES) framework for SFM, which consists of two kinds of screening rules, {\it{i.e.}}, \textbf{I}nactive \textbf{E}lements \textbf{S}creening (IES) and \textbf{A}ctive \textbf{E}lements \textbf{S}creening (AES). As we analyze above, the major challenge in developing IAES is the absence of KKT conditions. We bypass this obstacle by carefully studying the relationship between SFM and convex optimization, which can be regarded as another form of KKT conditions. We find that \ref{eqn:SFM} is closely related to a particular convex primal and dual problem pair \ref{eqn:Q-P} and \ref{eqn:Q-D} (see Section \ref{sec:basics}), that is, the minimizer of \ref{eqn:SFM} can be obtained from the positive components of the optimum of problem \ref{eqn:Q-P}. Hence, the proposed IAES identifies the active and inactive elements by estimating the lower and upper bounds of the components of the optimum of problem \ref{eqn:Q-P}. Thus, one of our major technical contributions is a novel  framework (Section \ref{sec:proposed-method})---developed by carefully studying the strong convexity of the corresponding primal and dual objective functions, the structure of the base polyhedra, and the optimality conditions of the SFM problem---for deriving accurate optimum estimation of problem \ref{eqn:Q-P}. We integrate IAES with the solver for problems \ref{eqn:Q-P} and \ref{eqn:Q-D}. As the solver goes on, and the estimation becomes more and more accurate, IAES can identify more and more elements. By fixing the active elements and removing the inactive ones, the problem size can be reduced gradually. IAES is safe in the sense that it would never sacrifice any accuracy on the final output. To the best of our knowledge, IAES is the first screening method in the domain of SFM or even combinatorial optimization. Moreover, compared with the screening methods for sparse models, an outstanding feature of IAES is that it has no theoretical limit in reducing the problem size. That is, we can finally reduce the problem size to zero, leading to substantial savings in the computational cost. The reason is that as the optimization proceeds, our estimation will be accurate enough to infer the  affiliations of all the elements with the optimizer $A^*$. While in sparse models, screening methods can never reduce the problem size to zero since the features (resp. samples) with nonzero coefficients in the primal (resp. dual) optimum can never be removed.  Experiments (see Section \ref{sec:experiment}) on both synthetic and real datasets demonstrate the significant speedups gained by IAES.  For the convenience of presentation, we postpone the detailed proofs of theoretical results in the main text to the supplementary materials. 

\textbf{Notations:} We consider a set $V=\{1, ..., p\}$, and denote its power set by $2^V$, which is composed of $2^p$ subsets of $V$. $|A|$ is the cardinality of a set $A$. $A\cup B$ and $A \cap B$ are the union and intersection of the sets $A$ and $B$, respectively. $A\subseteq B$ means that $A$ is a subset of $B$, potentially being equal to $B$. Moreover, for $\w \in \R^p$ and $\alpha \in \R$, we let $[\w]_k$ be the $k$-th component of $\w$ and $\{ \w \geq \alpha \}$ (resp. $\{\w> \alpha \}$) be the weak (resp. strong) $\alpha$-sup-level set of $\w$ defined as $\{k: k\in V, [\w]_k \geq \alpha \}$ (resp. $\{k: k\in V, [\w]_k > \alpha\}$). At last, for $\s \in \R^p$, we define a set function  by $\s(A) = \sum_{k\in A} [\s]_k$. 

\section{Basics and Motivations}\label{sec:basics}
This section is composed of two parts: a) briefly review some basics of submodular functions, SFM, and their relations with convex optimization; b) motivate our screening method IAES. 

The followings are the definitions of submodular function, submodular polyhedra and base polyhedra, which play an important role in submodular analysis. 
\begin{definition}\textup{[Submodular Function \cite{mccormick2005submodular}]} A set function $F: 2^V \rightarrow \R$ is submodular if and only if for all subsets $A, B\subseteq V$ we have:
	\begin{align}
	F(A)+F(B)\geq F(A\cup B) +F(A\cap B). \nonumber
	\end{align}
\end{definition}
\begin{definition}\textup{[Submodular and Base Polyhedra \cite{fujishige2005submodular}]} Let $F$ be a submodular function such that $F(\emptyset) = 0$. The submodular polyhedra $P(F)$ and the base polyhedra $B(F)$ are defined as:
	\begin{align}
	&P(F)\! =\! \{\s\! \in \R^p\!: \forall A \subseteq V, \s(A) \leq F(A) \}, \nonumber \\
	&B(F)\!=\!\{\s\! \in \R^p\!: \s(V) = F(V), \forall A \subseteq V, \s(A) \leq F(A)\}.\nonumber
	\end{align}
\end{definition}

Below we give the definition of Lov\'{a}sz extension, which works as the bridge that connects submodular functions and  convex functions.
\begin{definition}\textup{[Lov\'{a}sz Extension \cite{fujishige2005submodular}]} Given a set-function $F$ such that $F(\emptyset)=0$, the Lov\'{a}sz extension $f: \R^p \rightarrow \R$ is defined as follows: for $\w \in \R^p$, order the components in a decreasing order $[\w]_{j_1} \geq ... \geq [\w]_{j_p}$, and define $f(\w)$ through the equation below,
	\begin{align}
	&f(\w) = \sum_{k=1}^{p}[\w]_{j_k}\big(F(\{ j_1,...,j_k \})-F(\{ j_1,...,j_{k-1} \})\big). \nonumber
	\end{align}
\end{definition}

Lov\'{a}sz extension $f(\w)$ is convex if and only if $F$ is submodular (see \cite{fujishige2005submodular}).

%We focus on the generic submodular function minimization problem, which takes the form of 
%\begin{align}
%\min_{A \subseteq V} F(A), \tag{SFM} \label{eqn:SFM}
%\end{align}
%where $F(A)$ is a submodular function on $V$.

We focus on the generic submodular function minimization problem \ref{eqn:SFM} defined in Section \ref{sec:introduction} and denote its minimizer as $A^*$. To reveal the relationship between \ref{eqn:SFM} and convex optimization and finally motivate our method, we need the following theorems.  

\begin{theorem}\label{thm:dual-kkt} Let $\psi_1, ..., \psi_p$ be $p$ convex functions on $\R$, $\psi_1^*, ..., \psi_p^*$ be their Fenchel-conjugates \cite{borwein2010convex}, and $f$ be  the Lov\'{a}sz extension of a submodular function $F$. Denote the subgradient of $\psi_k(\cdot)$ by $\partial \psi_k(\cdot)$.  Then, the followings hold:\\
	$\rm{(i)}$ The problems below are dual of each other: 
	\begin{align}
	&\min_{ \w \in \R^p} f(\w) + \sum_{j=1}^{p}\psi_j([\w]_j), \tag{P} \label{eqn:primal} \\
	&\max_{\s\in B(F)} -\sum_{j=1}^{p} \psi_j^*(-[\s]_j). \tag{D} \label{eqn:dual}
	\end{align}
	$\rm{(ii)}$ The pair $(\w^*, \s^*)$ is optimal for problems (\ref{eqn:primal}) and (\ref{eqn:dual}) if and only if 
	\begin{align}
	\begin{cases}
	\textup{(a): } [\s]_k^* \in -\partial \psi_k([\w]_k^*), \forall k \in V, \\
	\textup{(b): } \w^* \in N_{B(F)}(\s^*) ,
	\end{cases} \tag{Opt} \label{opt-conditions}
	\end{align}		
	where $N_{B(F)}(\s^*)$ is the normal cone (see Chapter 2 of \cite{borwein2010convex}) of $B(F)$ at $\s^*$. 
\end{theorem}

When $\psi_j(\cdot)$ is differentiable, we consider a sequence of set optimization problems parameterized by $\alpha \in \R$:
\begin{align}
\min_{A\subseteq V} F(A) + \sum_{j\in A}\nabla \psi_j(\alpha), \tag{SFM'} \label{eqn:SFM'}
\end{align}
where $\nabla \psi_j(\cdot)$ is the gradient of $\psi_k(\cdot)$. The problem \ref{eqn:SFM'} has tight connections with the convex optimization problem \ref{eqn:primal} (see the theorem below).

\begin{theorem}\label{thm:convex-sfm}
	\textup{[Submodular function minimization from the proximal problem, Proposition 8.4 in \cite{bach2013learning}]} Under the same assumptions in Theorem \ref{thm:dual-kkt}, if $\psi_j(\cdot)$ is differentiable for all $j\in V$ and $\w^*$ is the unique minimizer of problem \ref{eqn:primal}, then for all $\alpha \in \R$, the minimal minimizer of problem \ref{eqn:SFM'} is $\{ \u > \alpha \}$ and the maximal minimizer is $\{\u \geq \alpha \}$, that is, for any minimizers $A^*_\alpha$ we have:
	\begin{align}
	\{ \w^* > \alpha \} \subseteq A^*_\alpha \subseteq \{ \w^* \geq \alpha \}. \label{eqn:estimation-A-alpha}
	\end{align}
\end{theorem}

By choosing $\psi_j(x)=\frac{1}{2}x^2$ and $\alpha = 0$ in \ref{eqn:SFM'}, combining Theorems \ref{thm:dual-kkt} and \ref{thm:convex-sfm}, we can see that  \ref{eqn:SFM} can be reduced to the following primal and dual problems, one is a quadratic optimization problem and the other is equivalent to finding the minimum norm point in the base polytope $B(F)$: 
\begin{align}
&\min_{ \w \in \R^p} P(\w) := f(\w) + \frac{1}{2}\|\w\|_2^2, \tag{Q-P} \label{eqn:Q-P}\\ 
&\max_{\s \in B(F)}{D(\s) := -\frac{1}{2}\|\s \|_2^2}. \tag{Q-D} \label{eqn:Q-D}
\end{align}

According to (\ref{eqn:estimation-A-alpha}), we can define two index sets:
\begin{align}
\calE := \{j \in V: [\w]_j^* > 0  \}, \mbox{ and }\calG := \{j \in V: [\w]_j^* < 0    \}, \nonumber 
\end{align}
which imply that 
\begin{align}
&\textup{(i): } j \in \calE \Rightarrow j \in A^*,\tag{R1} \label{eqn:R1} \\
&\textup{(ii): } j \in \calG \Rightarrow j \notin A^*. \tag{R2} \label{eqn:R2} 
\end{align}
We call the $j$-th element active if $j\in \calE$ and the ones in $\calG$ inactive. 

Suppose that we are given two subsets of $\calE$ and $\calG$, by rules \ref{eqn:R1} and \ref{eqn:R2}, we can see that many  affiliations between $A^*$ and the elements of $V$ can be deduced. Thus, we have less unknowns to solve in \ref{eqn:SFM} and its size can be dramatically reduced. We formalize this idea in Lemma \ref{lemma:scaled-problem}.

\begin{lemma}\label{lemma:scaled-problem}
	Given two subsets $\hcalG \subseteq \calG$ and $\hcalE \subseteq \calE$, the followings hold:
	
	\textup{(i):} $\hcalE \subseteq A^*$, and for all $j \in \hcalG$ we have $j\notin A^*$. 
	
	\textup{(ii):} The problem \ref{eqn:SFM} can be reduced to the following scaled problem:
	\begin{align}
	\min_{C \subseteq V/(\hcalE\cup \hcalG)} \hat{F}(C) := F(\hcalE \cup C) - F(\hcalE), \tag{scaled-SFM} \label{eqn:scaled-SFM}
	\end{align}
	which is also an SFM problem.
	
	\textup{(iii):} $A^*$ can be recovered by $A^* =\hcalE \cup C^*$, where $C^*$ is the minimizer of \ref{eqn:scaled-SFM}. 
\end{lemma}

Lemma \ref{lemma:scaled-problem} indicates that, if we can identify the active set $\hcalE$ and inactive set $\hcalG$, we only need to solve a scaled problem \ref{eqn:scaled-SFM}, which may have much smaller size than the original problem \ref{eqn:SFM}, to exactly recover the optimal solution $A^*$ without sacrificing any accuracy. 

However, since $\w^*$ is unknown, we cannot directly apply rules \ref{eqn:R1} and \ref{eqn:R2} to identify the active set $\hcalE$ and inactive set $\hcalG$. Inspired by the ideas in the gap safe screening methods (\cite{fercoq2015mind, NIPS2016_6405, shibagaki2016simultaneous}) for convex problems, we can first estimate the region $\calW$ that contains $\w^*$ and then relax the rules \ref{eqn:R1} and \ref{eqn:R2} to the practicable versions. Specifically, we first denote 
\begin{align}
&\hcalE := \{ j \in V: \min_{\w \in \calW} [\w]_j > 0  \}, \label{eqn:hcalE} \\
&\hcalG := \{ j \in V: \max_{\w \in \calW} [\w]_j < 0  \}.\label{eqn:hcalG}
\end{align}
It is obvious that $\hcalE \subseteq \calE$ and $\hcalG \subseteq \calG$. Hence, the rules \ref{eqn:R1} and \ref{eqn:R2} can be relaxed as follows:
\begin{align}
&\textup{(i): }  j \in \hcalE \Rightarrow j \in A^*,\tag{R1'} \label{eqn:R1'} \\
&\textup{(ii): }  j \in \hcalG \Rightarrow j \notin A^*. \tag{R2'} \label{eqn:R2'} 
\end{align}

In view of the rules \ref{eqn:R1'} and \ref{eqn:R2'}, we sketch the development of IAES as follows:\\
\textbf{Step 1:} Derive the estimation $\calW$ such that $\w^*\in\calW$.\\
\textbf{Step 2:} Develop IAES via deriving the detailed screening rules \ref{eqn:R1'} and \ref{eqn:R2'}.

\section{The Proposed Element Screening Method}\label{sec:proposed-method}
In this section, we first present the accurate optimum estimation by carefully studying the strong convexity of the functions $P(\w)$ and $D(\s)$, the optimality conditions of \ref{eqn:SFM} and its relationship with the convex problem pair (see Section \ref{sec:optimum-estimation}). Then, in Section \ref{sec:IES-AES}, we develop our inactive and active element screening rules IES and AES step by step. At last, in Section \ref{sec:IAES}, we develop the screening framework IAES by an alternating application of IES and AES.

\subsection{Optimum Estimation}\label{sec:optimum-estimation} 
Let $\hcalE$ and $\hcalG$ be the active and inactive sets identified by the previous IAES steps (before applying IAES for the first time, they are $\emptyset$). From Lemma \ref{lemma:scaled-problem}, we know that the problem \ref{eqn:SFM} then can be reduced to the following scaled problem:
\begin{align}
\min_{C \subseteq \hat{V}} \hat{F}(C) := F(\hcalE \cup C) - F(\hcalE), \nonumber
\end{align}
where $\hat{V}=V/(\hcalE\cup \hcalG)$. The second term $-F(\hcalE)$ at the right side of the equation above is added to make $\hat{F}(\emptyset)=0$. Thus, the corresponding problems \ref{eqn:Q-P} and \ref{eqn:Q-D} then become:
\begin{align}
&\min_{ \hat{\w} \in \R^{\hat{p}}} \hat{P}(\hat{\w}) := \hat{f}(\hat{\w}) + \frac{1}{2}\|\hat{\w}\|_2^2, \tag{Q-P'} \label{eqn:Q-P'}\\ 
&\max_{\hat{\s} \in B(\hat{F})}{\hat{D}(\hat{\s}) := -\frac{1}{2}\|\hat{\s}\|_2^2}, \tag{Q-D'} \label{eqn:Q-D'}
\end{align}
where $\hat{f}(\hat{\w})$ is the Lov\'{a}sz extension of $\hat{F}$ and $\hat{p}= |V/(\hcalE \cup \hcalG)|$. Now, we turn to estimate the minimizer $\hat{\w}^*$ of the problem \ref{eqn:Q-P'}. The result is presented in the theorem below. 

\begin{theorem}\label{thm:optimum-estimation-v1} For any $\hat{\w} \in dom \hat{P}(\hat{\w})$, $\hat{\s} \in B(\hat{F})$ and $C\subseteq \hat{V}$, we denote the dual gap as $G(\hat{\w}, \hat{\s})= \hat{P}(\hat{\w}) -\hat{D}(\hat{\s}) $, and then we have 
	\begin{align}
	\hat{\w}^* \in  \calW =   \calB \cap \Omega \cap \calP, \nonumber
	\end{align}	
	where $\calB = \Big\{\w: \|\w - \hat{\w} \| \leq \sqrt{2G(\hat{\w}, \hat{\s})} \Big \}$, $\Omega = \Big\{\w: \hat{F}(\hat{V}) - 2\hat{F}(C) \leq \|\w \|_1 \leq \|\hat{\s}\|_1 \Big \}$, and $\calP = \Big\{\w: \langle \w, \mathbf{1} \rangle = -\hat{F}(\hat{V}) \Big \}$. 
\end{theorem}
From the theorem above, we can see that the estimation $\calW$ is the intersection of three sets: the ball $\calB$, the $\ell_1$-norm equipped spherical shell $\Omega$ and the plane $\calP$. As the optimizer goes on, the dual gap $G(\hat{\w}, \hat{\s})$ becomes smaller, and $\hat{F}(\hat{V}) - 2\hat{F}(C)$ and $\|\hat{\w}\|_1$ would converge to $\|\hat{\w}^*\|_1$ (see Chapter 7 of \cite{bach2013learning}).  Thus, the volumes of $\calB$ and $\Omega$ become smaller and smaller during the optimization process, and the estimation $\calW$ would be more and more accurate. 

\subsection{Inactive and Active Element Screening}\label{sec:IES-AES}
We now turn to develop the screening rules IES and AES based on the estimation of the optimum $\hat{\w}^*$. 

From (\ref{eqn:hcalE}) and (\ref{eqn:hcalG}), we can see that, to develop the screening rules we need to solve two problems:
$\min_{\w\in \calW}[\w]_j$ and $\max_{\w\in \calW}[\w]_j$. 
However, since $\calW$ is highly non-convex and has a complex structure, it is very hard to solve these two problems efficiently. Hence, we rewrite the estimation $\calW$ as $\calW = (\calB \cap \calP)\cap (\calB \cap \Omega)$, and develop two different screening rules on $\calB \cap \calP$ and $\calB \cap \Omega$, respectively.
%
%\begin{theorem} \label{thm:IAES}
%	Given the active and inactive sets $\hcalE$ and $\hcalG$, which are identified in the previous IAES steps, then, 
%	
%	\textup{(i): }The active element screening rule takes the form of 
%	\begin{align}
%	\hat{\w}_j -\sqrt{2G(\hat{\w}, \hat{\s})} > 0 \Rightarrow j \in A^*, \forall j \in V/(\hcalE \cup \hcalG). \tag{AES} \label{rule:AES}
%	\end{align}
%	\textup{(ii): }The inactive element screening rule takes the form of 
%	\begin{align}
%	\hat{\w}_j +\sqrt{2G(\hat{\w}, \hat{\s})} < 0 \Rightarrow j \notin A^*, \forall j \in V/(\hcalE \cup \hcalG. \tag{IES} \label{rule:IES}
%	\end{align}
%	\textup{(iii): } The active and inactive sets $\hcalE$ and $\hcalG$ can be updated by 
%	\begin{align}
%	&\hcalE \leftarrow \hcalE \cup \Delta \hcalE, \label{eqn:update-E} \\
%	&\hcalF \leftarrow \hcalF \cup \Delta \hcalF, \label{eqn:update-G} 
%	\end{align}
%	where  $\Delta \hcalE $ and  $\Delta \hcalG$ are the newly identified active and inactive sets defined as 
%	\begin{align}
%	&\Delta \hcalE = \{ j \in V/(\hcalE \cup \hcalG) : \hat{\w}_j -\sqrt{2G(\hat{\w}, \hat{\s})} > 0 \}, \nonumber \\
%	&\Delta \hcalG = \{ j \in V/(\hcalE \cup \hcalG) : \hat{\w}_j +\sqrt{2G(\hat{\w}, \hat{\s})} < 0 \}. \nonumber 
%	\end{align}
%\end{theorem}

\subsubsection{Inactive and Active Element Screening based on $\calB \cap \calP$}
Given the estimation $\calB \cap \calP$, we derive the screening rules by solving the following problems：
\begin{align}
\min_{\w\in \calB\cap \calP}[\w]_j \mbox{ and }\max_{\w\in \calB \cap \calP}[\w]_j. \nonumber
\end{align}
We show that both of the two problems above admit closed-form solutions. 
\begin{lemma}\label{lemma:upper-lower-v1} Given the estimation ball $\calB$, the plane $\calP$ and the active and inactive sets $\hcalE$ and $\hcalG$, which are identified in the previous IAES steps, for all $j \in [\hat{p}]$ we denote
	\begin{align}
	&b_j =2\Big(\sum_{i\neq j} [\hat{\w}]_i+\hat{F}(\hat{V})-(\hat{p}-1)[\hat{\w}]_j\Big), \nonumber \\
	&c_j =\Big(\sum_{i\neq j} [\hat{\w}]_i + \hat{F}(\hat{V})\Big)^2- (\hat{p}-1)\Big(2G(\hat{\w}, \hat{\s})-[\hat{\w}]_j^2\Big).\nonumber
	\end{align}  
	Then the followings hold:
	\begin{align}
	&\textup{(i):} \min_{\w\in \calB\cap \calP}[\w]_j = [\w]_j^{\min} :=\frac{-b_j - \sqrt{b_j^2 - 4\hat{p}c_j}}{2\hat{p}},\nonumber \\
	&\textup{(ii):}\max_{\w\in \calB \cap \calP}[\w]_j = [\w]_j^{\max} :=\frac{-b_j + \sqrt{b_j^2 - 4\hat{p}c_j}}{2\hat{p}}. \nonumber
	\end{align}
\end{lemma}
%
%\begin{lemma}\label{lemma:upper-lower-v-tmp} Given the estimation ball $\calB$, the plane $\calP$ and the active and inactive sets $\hcalE$ and $\hcalG$, which are identified in the previous IAES steps, for all $\forall j \in \hat{p}$, we denote
%	\begin{align}
%	&b_j =2\Big(\sum_{i\neq j} [\hat{\s}]_i-\hat{F}(\hat{V})-(\hat{p}-1)[\hat{\s}]_j\Big), \nonumber \\
%	&c_j =\Big(\sum_{i\neq j} [\hat{\s}]_i - \hat{F}(\hat{V})\Big)^2- (\hat{p}-1)\Big(2G(\hat{\w}, \hat{\s})-[\hat{\s}]_j^2\Big),\nonumber
%	\end{align}  
%	then the followings hold:
%	\begin{align}
%	&\textup{(i):} \min_{\s\in \calB\cap \calP}[\s]_j = [\s]_j^{\min} :=\frac{-b_j - \sqrt{b_j^2 - 4\hat{p}c_j}}{2\hat{p}}.\nonumber \\
%	&\textup{(ii):}\max_{\s\in \calB \cap \calP}[\s]_j = [\s]_j^{\max} :=\frac{-b_j + \sqrt{b_j^2 - 4\hat{p}c_j}}{2\hat{p}}. \nonumber
%	\end{align}
%\end{lemma}

We are now ready to present the active and inactive screening rules \ref{rule:AES-v1} and \ref{rule:IES-v1}. 
\begin{theorem} \label{thm:IAES-v1}
	Given the active and inactive sets $\hcalE$ and $\hcalG$, which are identified in the previous IAES steps, we have
	
	\textup{(i): }The active element screening rule takes the form of 
	\begin{align}
	[\w]_j^{\min}> 0 \Rightarrow j \in A^*, \forall j \in V/(\hcalE \cup \hcalG). \tag{AES-1} \label{rule:AES-v1}
	\end{align}
	\textup{(ii): }The inactive element screening rule takes the form of 
	\begin{align}
	[\w]_j^{\max}< 0 \Rightarrow j \notin A^*, \forall j \in V/(\hcalE \cup \hcalG). \tag{IES-1} \label{rule:IES-v1}
	\end{align}
	\textup{(iii):} The active and inactive sets $\hcalE$ and $\hcalG$ can be updated by 
	\begin{align}
	&\hcalE \leftarrow \hcalE \cup \Delta \hcalE, \label{eqn:update-E-v1} \\
	&\hcalG \leftarrow \hcalG \cup \Delta \hcalG, \label{eqn:update-G-v1} 
	\end{align}
	where  $\Delta \hcalE $ and  $\Delta \hcalG$ are the newly identified active and inactive sets defined as 
	\begin{align}
	&\Delta \hcalE := \{ j \in V/(\hcalE \cup \hcalG) : [\w]_j^{\min} > 0 \}, \nonumber \\
	&\Delta \hcalG := \{ j \in V/(\hcalE \cup \hcalG) : [\w]_j^{\max}< 0 \}. \nonumber 
	\end{align}
\end{theorem}

From the theorem above, we can see that our rules \ref{rule:AES-v1} and \ref{rule:IES-v1} are safe in the sense that the detected elements are guaranteed to be included or excluded in $A^*$. 
%\begin{theorem} \label{thm:IAES-v1-temp}
%	Given the active and inactive sets $\hcalE$ and $\hcalG$, which are identified in the previous IAES steps, then,
%	
%	\textup{(i): }The active element screening rule takes the form of 
%	\begin{align}
%	[\s]_j^{\max}< 0 \Rightarrow j \in A^*, \forall j \in V/(\hcalE \cup \hcalG). \tag{AES-1} \label{rule:AES-v1-temp}
%	\end{align}
%	\textup{(ii): }The inactive element screening rule takes the form of 
%	\begin{align}
%	[\w]_j^{\min}> 0 \Rightarrow j \notin A^*, \forall j \in V/(\hcalE \cup \hcalG. \tag{IES-1} \label{rule:IES-v1-temp}
%	\end{align}
%\end{theorem}
\subsubsection{Inactive and Active Element Screening based on $\calB \cap \Omega$}
We now derive the second screening rule pair based on the estimation $\calB \cap \Omega$. 

Due to the high non-convexity and complex structure of $\calB \cap \Omega$, directly solving problems $\min_{\w\in \calB\cap \Omega}[\w]_j$ and $\max_{\w\in \calB \cap \Omega}[\w]_j$ is time consuming. Notice that, to derive IAS and IES, we only need to judge whether the inequalities $\min_{\w\in \calB\cap \Omega}[\w]_j > 0$ and $\max_{\w\in \calB\cap \Omega}[\w]_j < 0$ are satisfied or not, instead of calculating $\min_{\w\in \calB\cap \Omega}[\w]_j$ and $\max_{\w\in \calB\cap \Omega}[\w]_j$. Hence, we only need to infer the hypotheses $\big \{ \w: \w\in \calB, [\w]_j \leq 0 \big \} \cap \Omega = \emptyset$ and $\big \{ \w: \w\in \calB, [\w]_j \geq 0 \big \} \cap \Omega = \emptyset$ are true or false. Thus, from the formulation of $\Omega$ (see Theorem \ref{thm:optimum-estimation-v1}), the problems boil down to calculating the minimum and the maximum of $\|\w\|_1$ with $\big\{ \w: \w\in \calB, [\w]_j \geq 0 \big \}$ or $\big \{ \w: \w\in \calB, [\w]_j \leq 0 \big \}$, which admit closed-form solutions. The results are presented in the lemma below.

\begin{lemma}\label{lemma:upper-lower-v2}
	Given the estimation ball $\calB$ and the active and inactive sets $\hcalE$ and $\hcalG$, which are identified in the previous IAES steps, then the followings hold:
	
	\textup{(i): } $\forall j \in \hat{p}$, if $|[\hat{\w}]_j|> \sqrt{2G(\hat{\w}, \hat{\s})}$, then the element $j$ can be identified by rule \ref{rule:AES-v1} or \ref{rule:IES-v1} to be active or inactive.\\
	\textup{(ii): } $\forall j \in \hat{p}$, if $0 < [\hat{\w}]_j \leq  \sqrt{2G(\hat{\w}, \hat{\s})}$, we have
	\begin{align}
	& \min_{\w\in \calB, [\w]_j \leq 0} \|\w\|_1 < \|\hat{\w}\|_1, \nonumber \\
	&\max_{\w\in \calB, [\w]_j \leq 0} \|\w\|_1 =\!\! \begin{cases}
	\!\|\hat{\w}\|_1\!-\!\!2[\hat{\w}]_j \!+\!\sqrt{2\hat{p}G(\hat{\w}\!,\! \hat{\s})}, \mbox{if }\!  [\hat{\w}]_j \!-\!\! \sqrt{\frac{2G(\hat{\w}, \hat{\s})}{\hat{p}}}\!\! <\! 0,\\
	\|\hat{\w}\|_1\!-\! [\hat{\w}]_j \!+\!   \sqrt{\hat{p}\!-\!1} \sqrt{2G(\hat{\w}\!,\! \hat{\s}) \!-\!  [\hat{\w}]_j^2}, \mbox{otherwise}.\nonumber
	\end{cases} 
	\end{align} 
	\textup{(iii): } $\forall j \in \hat{p}$, if $-\sqrt{2G(\hat{\w}, \hat{\s})} \leq  [\hat{\w}]_j < 0$, we have 
	\begin{align}
	& \min_{\w\in \calB, [\w]_j \geq 0} \|\w\|_1 < \|\hat{\w}\|_1, \nonumber \\
	&\max_{\w\in \calB, [\w]_j \geq 0} \|\w\|_1 = \!\! \begin{cases}
	\!\|\hat{\w}\|_1\!+\!\!2[\hat{\w}]_j \!+\!\sqrt{2\hat{p}G(\hat{\w}\!,\! \hat{\s})}, \mbox{if }\! [\hat{\w}]_j\! +\!\! \sqrt{\frac{2G(\hat{\w}, \hat{\s})}{\hat{p}}}\!\! >\! 0,\\
	\|\hat{\w}\|_1\!+\! [\hat{\w}]_j \!+\!   \sqrt{\hat{p}\!-\!1} \sqrt{2G(\hat{\w}\!,\! \hat{\s})\! -\! [\hat{\w}]_j^2}, \mbox{otherwise}.\nonumber 
	\end{cases} 
	\end{align}
\end{lemma}

We are now ready to present the second active and inactive screening rule pair \ref{rule:AES-v2} and \ref{rule:IES-v2}. From the lemma above, we can see that the element $j$ with $|[\hat{\w}]_j |>\sqrt{2G(\hat{\w}, \hat{\s})}$ can be screened by rules \ref{rule:AES-v1} and \ref{rule:IES-v1}. Hence, we now only need to consider the cases when $|[\hat{\w}]_j |\leq \sqrt{2G(\hat{\w}, \hat{\s})}$.
\begin{theorem} \label{thm:IAES-v2}
	Given a set $C\subseteq \hat{V}$ and the active and inactive sets $\hcalE$ and $\hcalG$ identified in the previous IAES steps, then,
	
	\textup{(i): }The active element screening rule takes the form of 
	\begin{align}
	\begin{cases}
	0 < [\hat{\w}]_j \leq  \sqrt{2G(\hat{\w}, \hat{\s})} \nonumber \\
	\max_{\w\in \calB, [\w]_j \leq 0} \|\w\|_1 < \hat{F}(\hat{V})-2\hat{F}(C) \nonumber
	\end{cases} \Rightarrow  j \in A^*, \forall j \in V/(\hcalE \cup \hcalG). \tag{AES-2} \label{rule:AES-v2}
	\end{align}
	\textup{(ii): }The inactive element screening rule takes the form of 
	\begin{align}
	\begin{cases}
	-\sqrt{2G(\hat{\w}, \hat{\s})} \leq [\hat{\w}]_j < 0 \nonumber \\
	\max_{\w\in \calB, [\w]_j \geq 0} \|\w\|_1 < \hat{F}(\hat{V})-2\hat{F}(C) \nonumber
	\end{cases} \Rightarrow j \notin A^*, \forall j \in V/(\hcalE \cup \hcalG). \tag{IES-2} \label{rule:IES-v2}
	\end{align}
	\textup{(iii):} The active and inactive sets $\hcalE$ and $\hcalG$ can be updated by 
	\begin{align}
	&\hcalE \leftarrow \hcalE \cup \Delta \hcalE, \label{eqn:update-E-v2} \\
	&\hcalG \leftarrow \hcalG \cup \Delta \hcalG, \label{eqn:update-G-v2} 
	\end{align}
	where  $\Delta \hcalE $ and  $\Delta \hcalG$ are the newly identified active and inactive sets defined as 
	\begin{align}
	\Delta \hcalE := &\Big\{ j \in V/(\hcalE \cup \hcalG) :
	0 < [\hat{\w}]_j \leq  \sqrt{2G(\hat{\w}, \hat{\s})}, \max_{\w\in \calB, [\w]_j \leq 0} \|\w\|_1 < \hat{F}(\hat{V})-2\hat{F}(C) \Big \}, \nonumber\\
	\Delta \hcalG := &\Big\{ j \in V/(\hcalE \cup \hcalG) :
	-\sqrt{2G(\hat{\w}, \hat{\s})} \leq [\hat{\w}]_j < 0, \max_{\w\in \calB, [\w]_j \geq 0} \|\w\|_1 < \hat{F}(\hat{V})-2\hat{F}(C) \Big \}. \nonumber
	\end{align}
\end{theorem}
Theorem \ref{thm:IAES-v2} verifies the safety of \ref{rule:AES-v2} and \ref{rule:IES-v2}. 

\subsection{The Proposed IAES Framework by An Alternating Execution of AES and IES} \label{sec:IAES}
To reinforce the capability of the proposed screening rules, we develop a novel framework IAES in Algorithm \ref{alg:IAES}, which applies the active element screening rules (\ref{rule:AES-v1} and \ref{rule:AES-v2}) and the inactive element screening rules (\ref{rule:IES-v1} and \ref{rule:IES-v2}) in an alternating manner during the optimization process. Specifically, we integrate our screening rules \ref{rule:AES-v1}, \ref{rule:AES-v2}, \ref{rule:IES-v1} and \ref{rule:IES-v2} with the optimization algorithm $\calA$ for the problems \ref{eqn:Q-P'} and \ref{eqn:Q-D'}. During the optimization process, we trigger the screening rules \ref{rule:AES-v1}, \ref{rule:AES-v2}, \ref{rule:IES-v1} and \ref{rule:IES-v2} every time when the dual gap is $1-\rho$ times smaller than itself in the last triggering of IAES. As the solver $\calA$ goes on, the volumes of $\Omega$ and $\calB$ would decrease to zeros quickly, IAES can thus identify more and more inactive and active elements. 

Compared with the existing screening methods for convex sparse models, an appealing feature of IAES is that it has no theoretical limit in identifying the inactive and active elements and reducing the problem size. The reason is that, in convex sparse models, screening models can never rule out the features and samples whose corresponding coefficients in the optimal solution are nonzero. While in our case, as the optimizer $\calA$ goes on, our estimation will be accurate enough for us to infer the affiliation of each element with $A^*$. Hence, we can finally identify all the inactive and active elements and the problem size can be reduced to zero. This nice feature can lead to significant speedups in the computation time.    

\begin{algorithm}[tb]\caption{Inactive and Active Element Screening}
	\begin{algorithmic}[1]
		\STATE {\bfseries Input: } an optimization algorithm $\calA$ for problems (\ref{eqn:Q-P'}) and (\ref{eqn:Q-D'}), $\epsilon > 0, 0< \rho < 1$. 
		\STATE {\bfseries Initialize:} $\hcalE =\hcalG =  \emptyset, C = \emptyset, g=\infty $, choose $\hat{\s} \in B(F) $ and $\hat{\w} = -\hat{\s}$.
		\REPEAT
		\STATE Run $\calA$ on problems (\ref{eqn:Q-P'}) and (\ref{eqn:Q-D'}) to update $\hat{\w}$, $\hat{\s}$ and $C$. 
		\IF{ dual gap $G(\hat{\w},\hat{\s})<\rho g$}
		\STATE Run the active element screening rules \ref{rule:AES-v1} and \ref{rule:AES-v2} based on $(\hat{\w},\hat{\s})$ and $C$. 
		\STATE Update the active set $\hcalE$ by (\ref{eqn:update-E-v1}) and (\ref{eqn:update-E-v2}).
		\STATE Run the inactive element screening rules \ref{rule:IES-v1} and \ref{rule:IES-v2} based on $(\hat{\w},\hat{\s})$ and $C$. 
		\STATE Update the inactive set $\hcalG$ by (\ref{eqn:update-G-v1}) and (\ref{eqn:update-G-v2}). 
		\IF{$V/(\hcalE \cup \hcalG)=\emptyset$}
		\STATE {\bfseries Return: } $\hcalE$.
		\ELSE
		\STATE Update $\hat{F}$, \ref{eqn:Q-P'}, \ref{eqn:Q-D'} according to $\hcalE$ and $\hcalG$. 
		\STATE Update $\hat{\w}$ and $\hat{\s}$ by:
		\begin{align}
		& \hat{\w} \leftarrow [\hat{\w}]_{V/(\hcalE \cup \hcalG)},\nonumber \\
		& \hat{\s} \leftarrow \arg \max_{\s \in B(\hat{F})} \langle \hat{\w}, \s \rangle.  \nonumber 
		\end{align}
		\STATE Update $g \leftarrow G(\hat{\w},\hat{\s})$.
		\ENDIF
		\ENDIF
		\UNTIL {$G(\hat{\w}, \hat{\s}) < \epsilon$.}
		\STATE {\bfseries Return:} $\hcalE \cup \{ \hat{\w}>0 \}$. 
	\end{algorithmic}\label{alg:IAES}
\end{algorithm}
\begin{remark}
	The set $C$ in Algorithm \ref{alg:IAES} is updated by choosing one of the super-level sets of $\hat{\w}$ with the smallest value $\hat{F}(C)$. It is free to obtain it. The reason is that most of the existing methods $\calA$ for the problems \ref{eqn:Q-P'} and \ref{eqn:Q-D'} need to calculate $\hat{f}(\hat{\w})$ in each iteration, in which they need to calculate the value $\hat{F}$ at all of the super-level sets of $\hat{\w}$ (see the greedy algorithm in \cite{bach2013learning} for details). 
\end{remark}

\begin{remark}
	The algorithm $\calA$ can be all the methods for the problems \ref{eqn:Q-P'} and \ref{eqn:Q-D'}, such as  minimum-norm point
	algorithm \cite{wolfe1976finding} and conditional gradient descent \cite{dunn1978conditional}. Although some algorithms only update $\s$, in IAES, we can update $\w$ in each iteration by letting $\w=-\s$ and refining it by the algorithm named pool adjacent violators \cite{best1990active}.
\end{remark}

%\begin{remark}\label{remark:screening-time}
%	Calculating the dual gap $G(\hat{\w}, \hat{\s})$ is very cheap, we do not need to calculate the expensive component $\hat{f}(\hat{\w})$ since most of the existing methods for the problems (\ref{eqn:Q-P'}) and (\ref{eqn:Q-D'}) would calculate it in each iteration. 
%\end{remark}
\begin{remark}
	Due to Lemma \ref{lemma:scaled-problem} and the  safety of \ref{rule:AES-v1}, 
	\ref{rule:AES-v2}, \ref{rule:IES-v1} and \ref{rule:IES-v2},
	we can see that IAES would never sacrifice any accuracy.
\end{remark}
\begin{remark}
	Although step 14 in Algorithm \ref{alg:IAES} may  increase the dual gap slightly, it is worthwhile because of the reduced problem size. This is verified by the speedups gained by IAES in the experiments.   
\end{remark}
\begin{remark}
	The parameter $\rho$ in Algorithm \ref{alg:IAES} controls the frequency how often we trigger IAES. The larger value, the higher frequency to trigger IAES but more computational time consumed by IAES. In our experiment, we set $\rho = 0.5$ and it achieves a good performance.
\end{remark}
\section{Experiments}\label{sec:experiment}
We evaluate IAES through numerical experiments on both synthetic and real datasets by two measurements. The first one is the rejection ratios of IAES over iterations: $\frac{m_i+n_i}{m^*+n^*}$, where $m_i$ and $n_i$ are the numbers of the active and inactive elements identified by IAES after the $i$-th iteration, and $m^*$ and $n^*$ are the numbers of the active and inactive elements in $A^*$. We notice that in our experiments $m^* + n^*=p$, so the rejection ratio presents the problem size reduced by IAES. The second measurement is speedup, i.e., the ratio of the running times of the solver without IAES and with IAES. We set the accuracy $\epsilon$ to be $10^{-6}$. 

Recall that, IAES can be integrated with all the solvers for the problems \ref{eqn:Q-P} and \ref{eqn:Q-D}. In this experiment, we  use one of the most widely used algorithm  minimum-norm point
algorithm (MinNorm) \cite{wolfe1976finding} as the solver. The function $F(A)$ varies according to the datasets, whose detailed definitions will be given in subsequent sections.

We write the code in Matlab and perform all the computations on a single core
of Intel(R) Core(TM) i7-5930K 3.50GHz, 32GB MEM.

\subsection{Experiments on Synthetic Datasets}

\begin{figure}[htb!]
	\begin{center}
		\includegraphics[scale = 0.41]{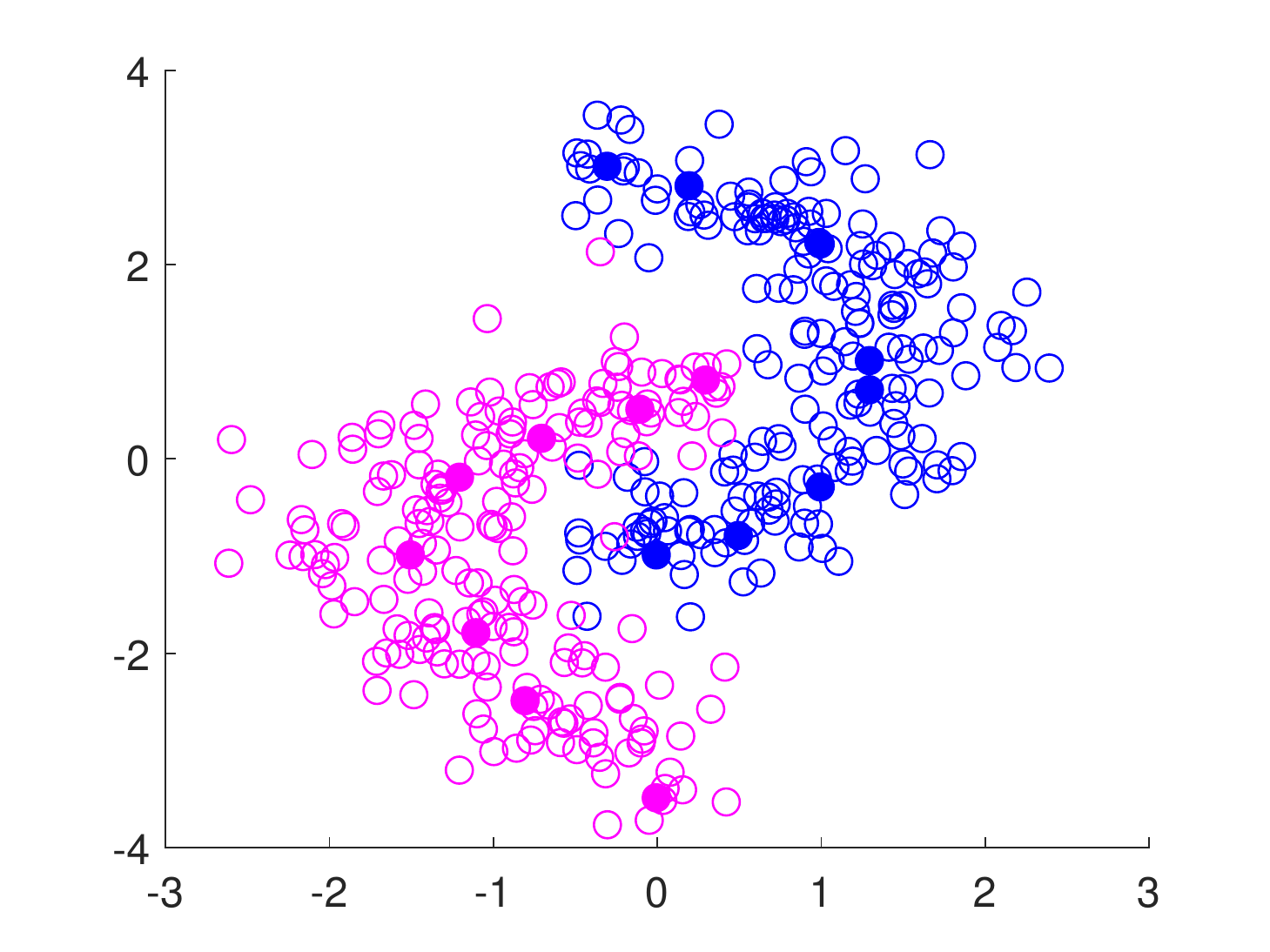}
		\caption{Two-moons dataset with 400 samples. The data points in magenta and blue are sampled from two different distributions. The filled dots are the labeled data points.}
		\label{fig:data-example}
	\end{center}
\end{figure}

We perform experiments on a synthetic dataset named two-moons with different sample sizes (see Figure \ref{fig:data-example} for an example). All the data points are sampled from two different semicircles. Specifically, each point can be represented as $\x = \c_i + \gamma *[\cos(\theta_i), \sin(\theta_i)]$, where $i= 1,2$ stands for the two semicircles, $\c_1 = [-0.5,1], \c_2=[0.5,-1]$, $\gamma$ is generated from a normal distribution $N(2,0.5^2)$, and  $\theta_1$ and $\theta_2$ are sampled from two uniform  distributions $[-\frac{\pi}{2},\frac{\pi}{2}]$ and $[\frac{\pi}{2},\frac{3\pi}{2}]$, respectively.  We first sample $p$ data points from these two semicircles with equal probability. Then, we randomly choose $p_0=16$ samples and label each of them as positive if it is from the first semicircle and otherwise label it as negative. We generate five datasets by varying the sample size $p$ in $[200,400,600,800, 1000]$. We perform semi-supervised clustering on each dataset and the objective function $F(A)$ is defined as:
\begin{align}
F(A) = I(f_A, f_{V/A}) - \sum_{j \in A} \log \eta_j - \sum_{j \in V/A} \log (1-\eta_j), \nonumber
\end{align}
where $I(f_A, f_{V/A})$ is the mutual information between two Gaussian processes with a Gaussian kernel $k(x,y) = \exp(-\alpha \|\x-\y\|^2), \alpha = 1.5$, and $\eta_j \in \{0,1\}$ if $j$ is labeled and otherwise $\eta_j=\frac{1}{2}$ (see Chapter 6 of \cite{bach2013learning} for more details). The kernel matrix is dense with the size $p \times p$, leading to a big computational cost when $p$ is large.

\begin{figure*}[htb!]
	\begin{center}
		\subfigure[$p$=200]{\includegraphics[scale=0.20]{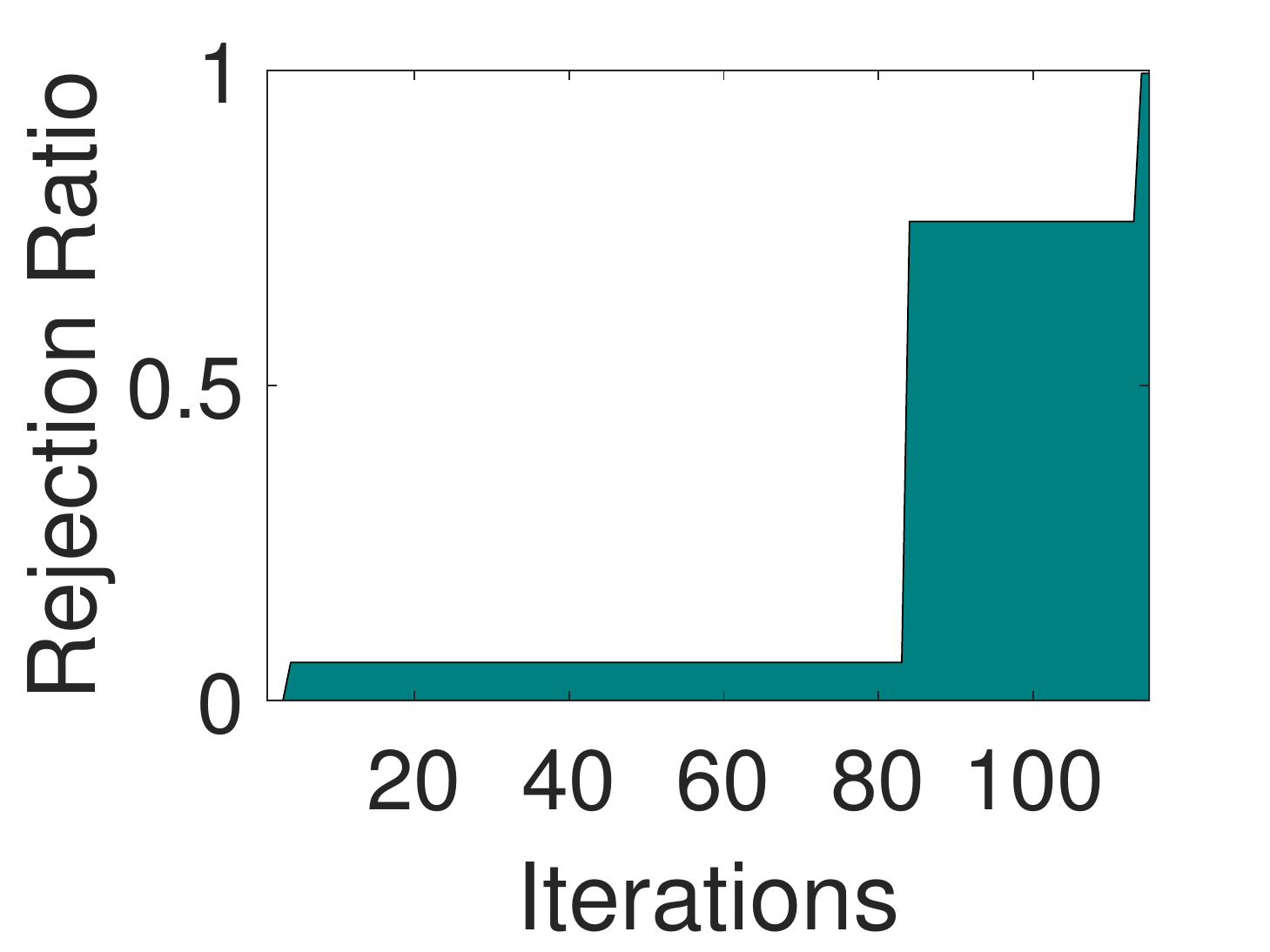}}
		\subfigure[$p$=400]{\includegraphics[scale=0.20]{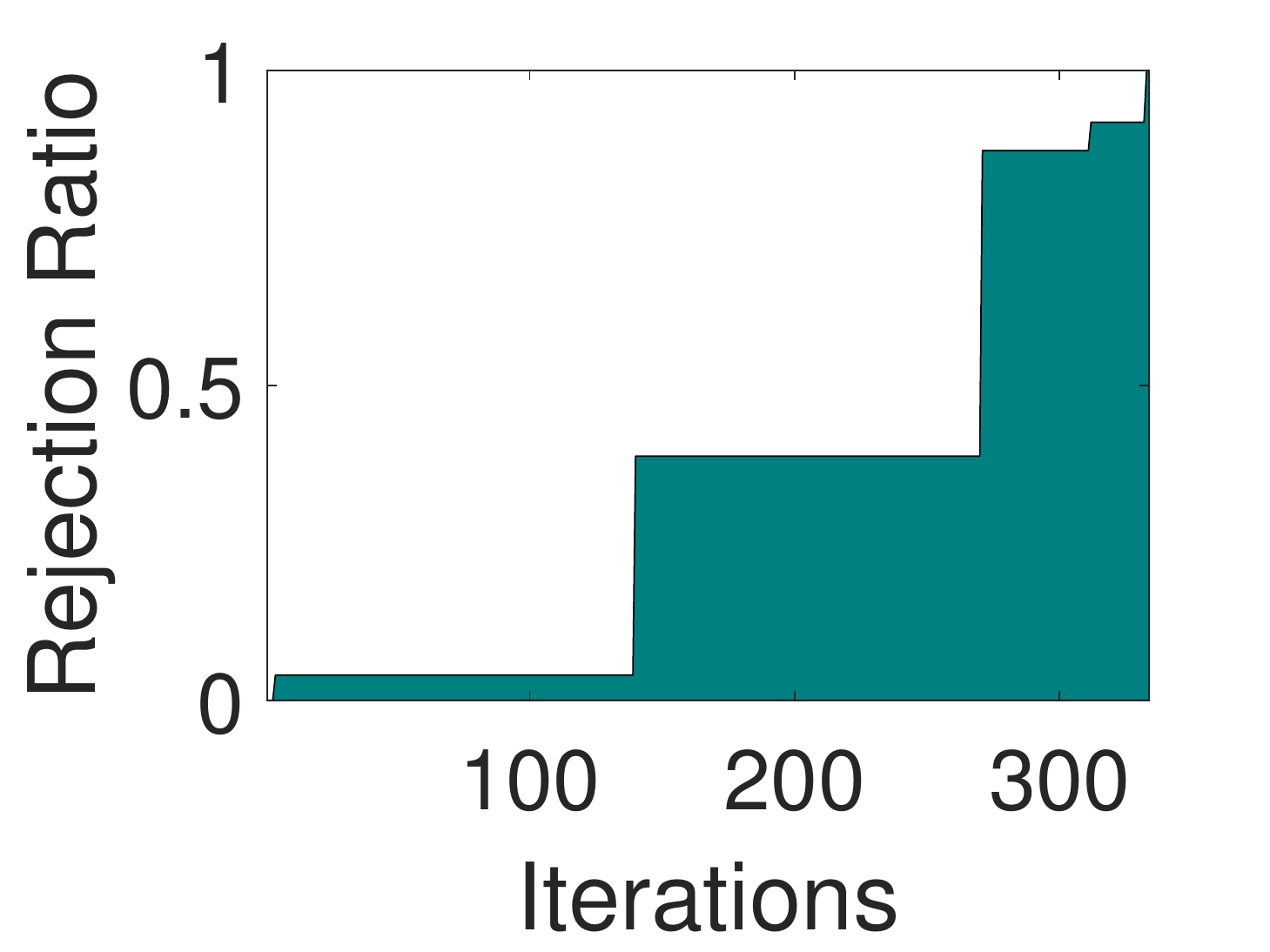}}
		\subfigure[$p$=600]{\includegraphics[scale=0.20]{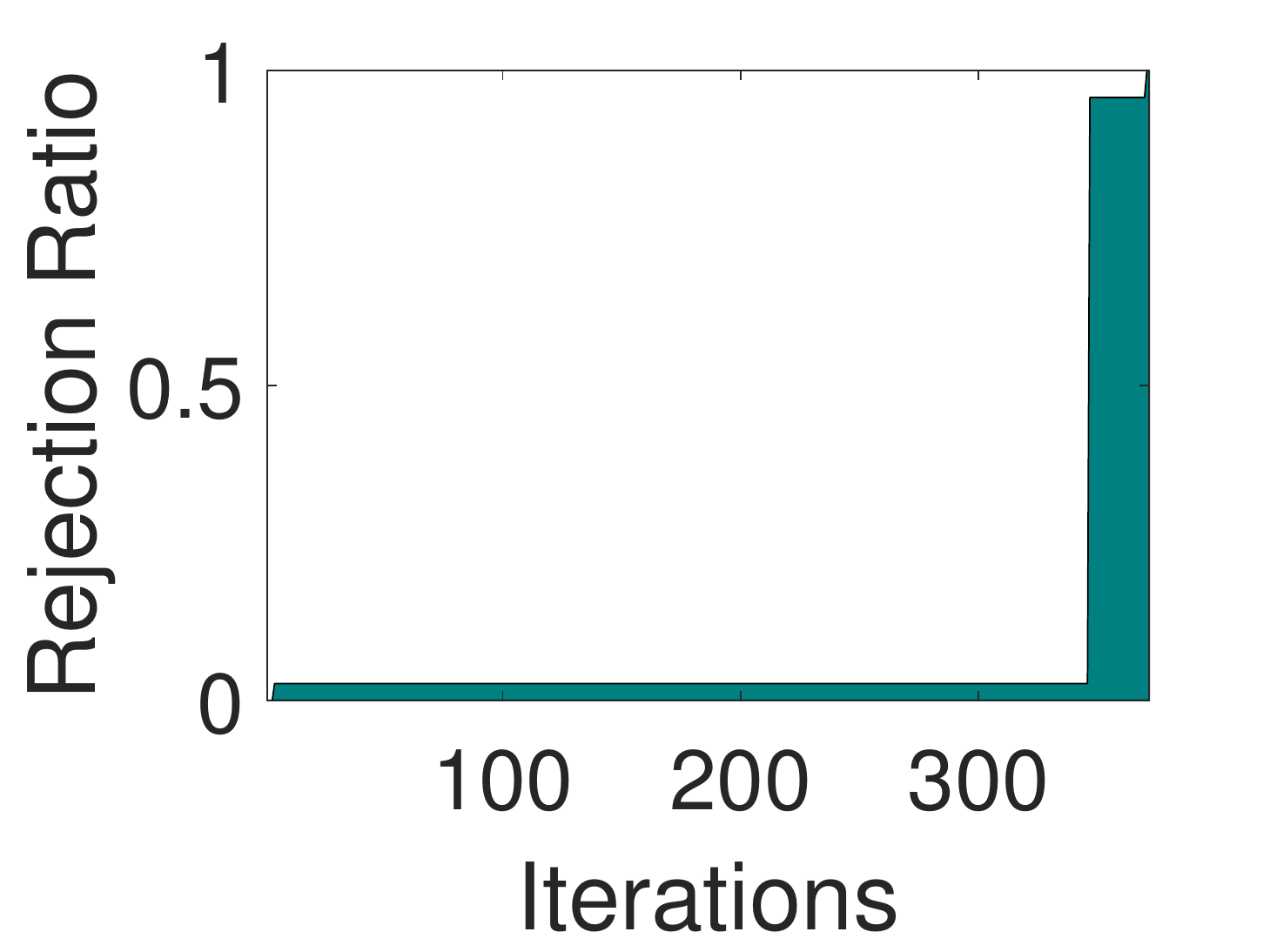}}
		\subfigure[$p$=800]{\includegraphics[scale=0.20]{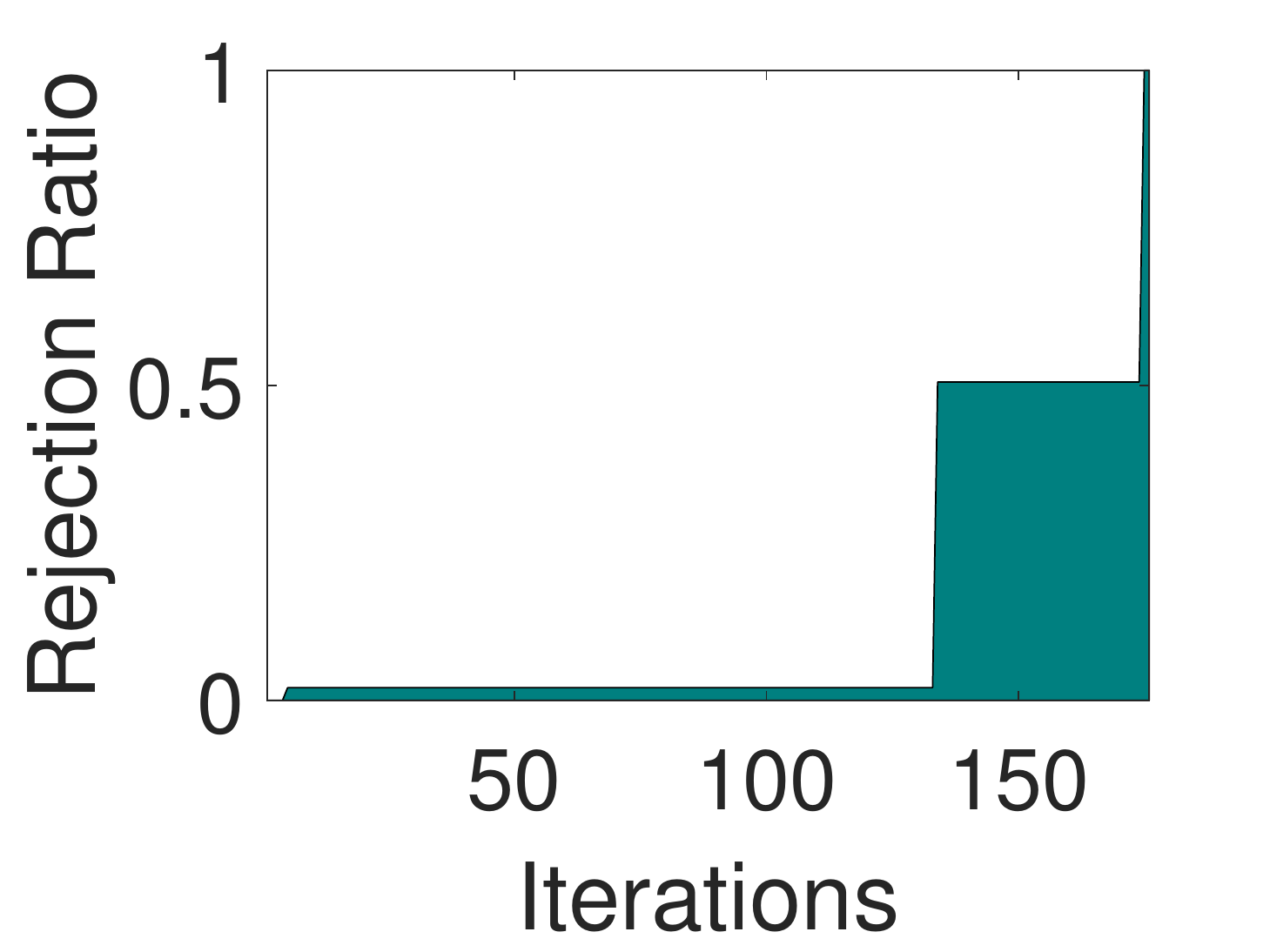}}
		\subfigure[$p$=1000]{\includegraphics[scale=0.20]{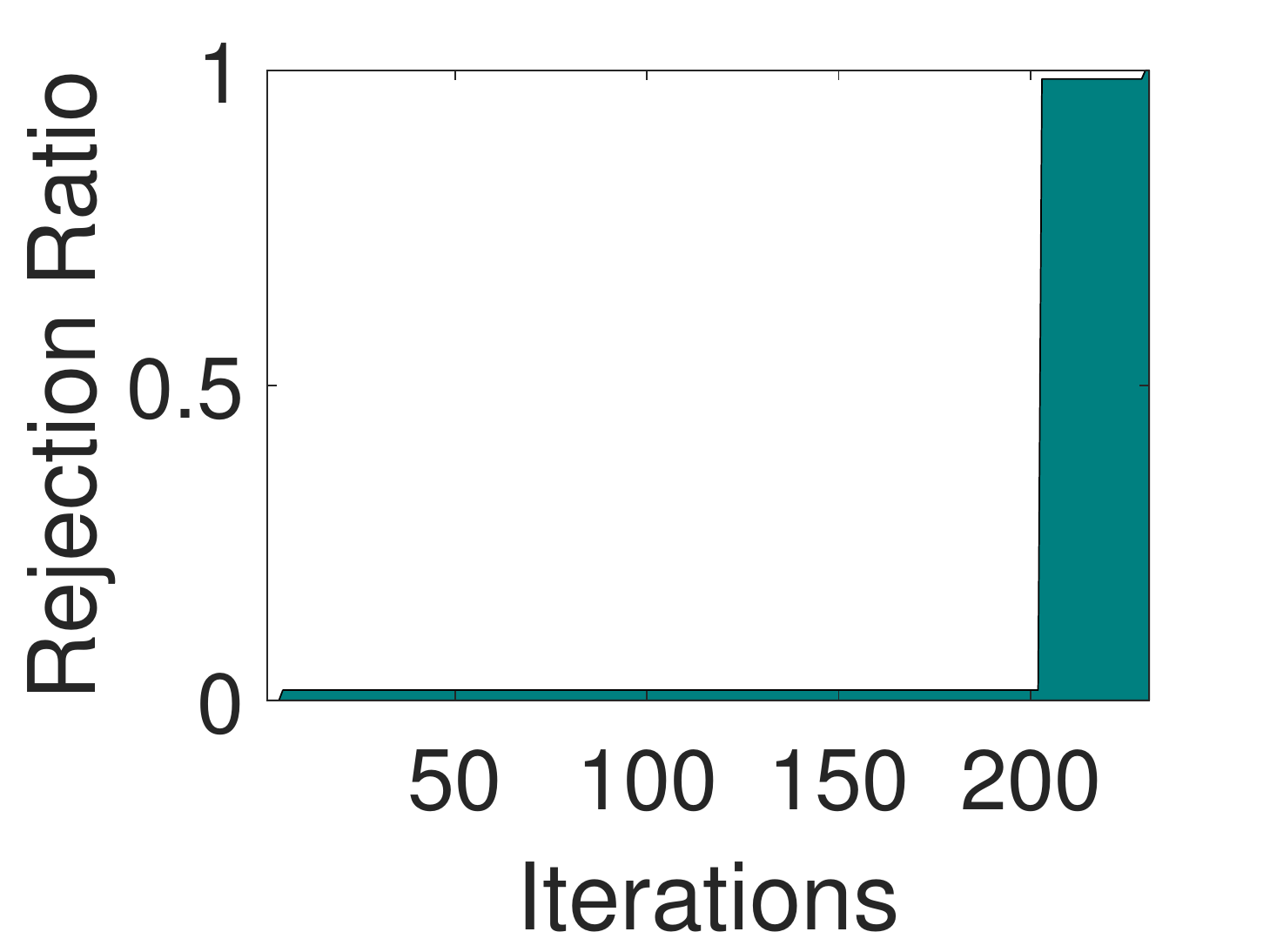}}
		\caption{Rejection ratios of IAES over the iterations on two-moons.}
		\label{fig:reject-ratios-syn}
	\end{center}
\end{figure*}

Figure \ref{fig:reject-ratios-syn} displays the rejection ratios of IAES on two-moons. We can see that IAES can find the active and inactive elements incrementally  during the optimization process. It can finally identify almost all of the elements and reduce the problem size to nearly zero in no more than 400 iterations, which is consistent with our theoretical analysis in Section \ref{sec:IAES}.

\begin{figure*}[htb!]
	\begin{center}
		\subfigure[after $4$-th iteration]{\includegraphics[scale=0.20]{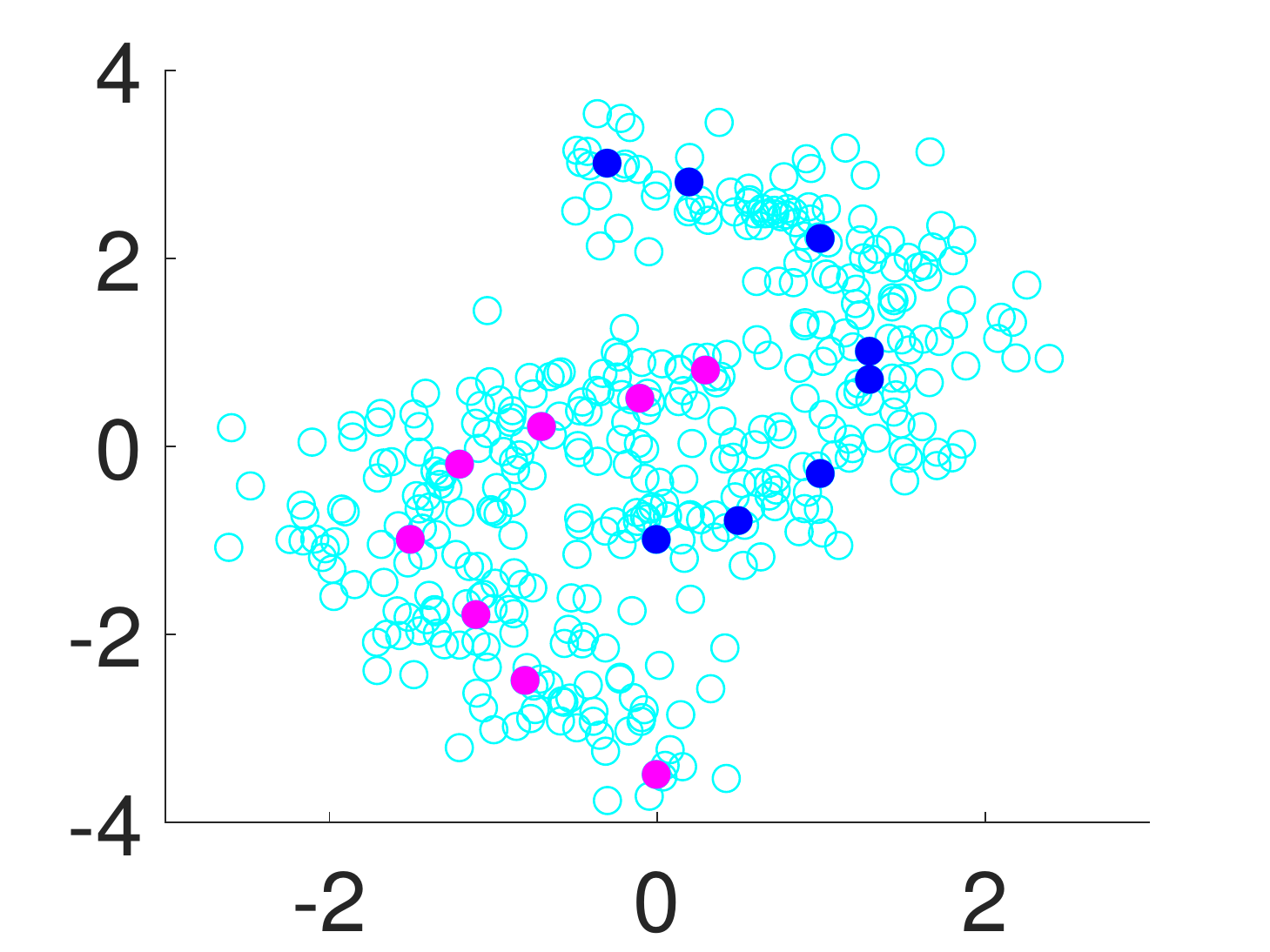}}
		\subfigure[after $140$-th iteration]{\includegraphics[scale=0.20]{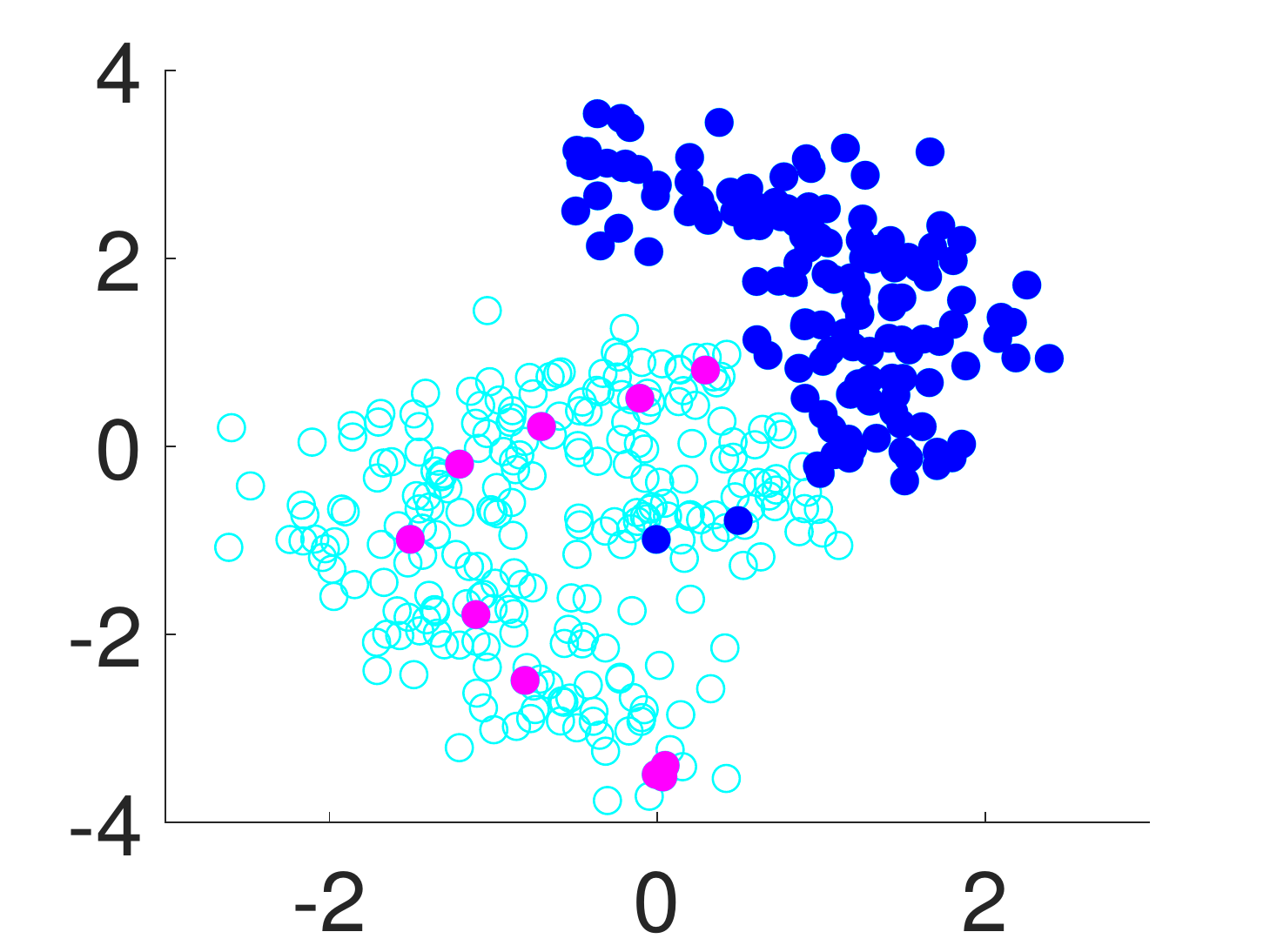}}
		\subfigure[after $271$-th iteration]{\includegraphics[scale=0.20]{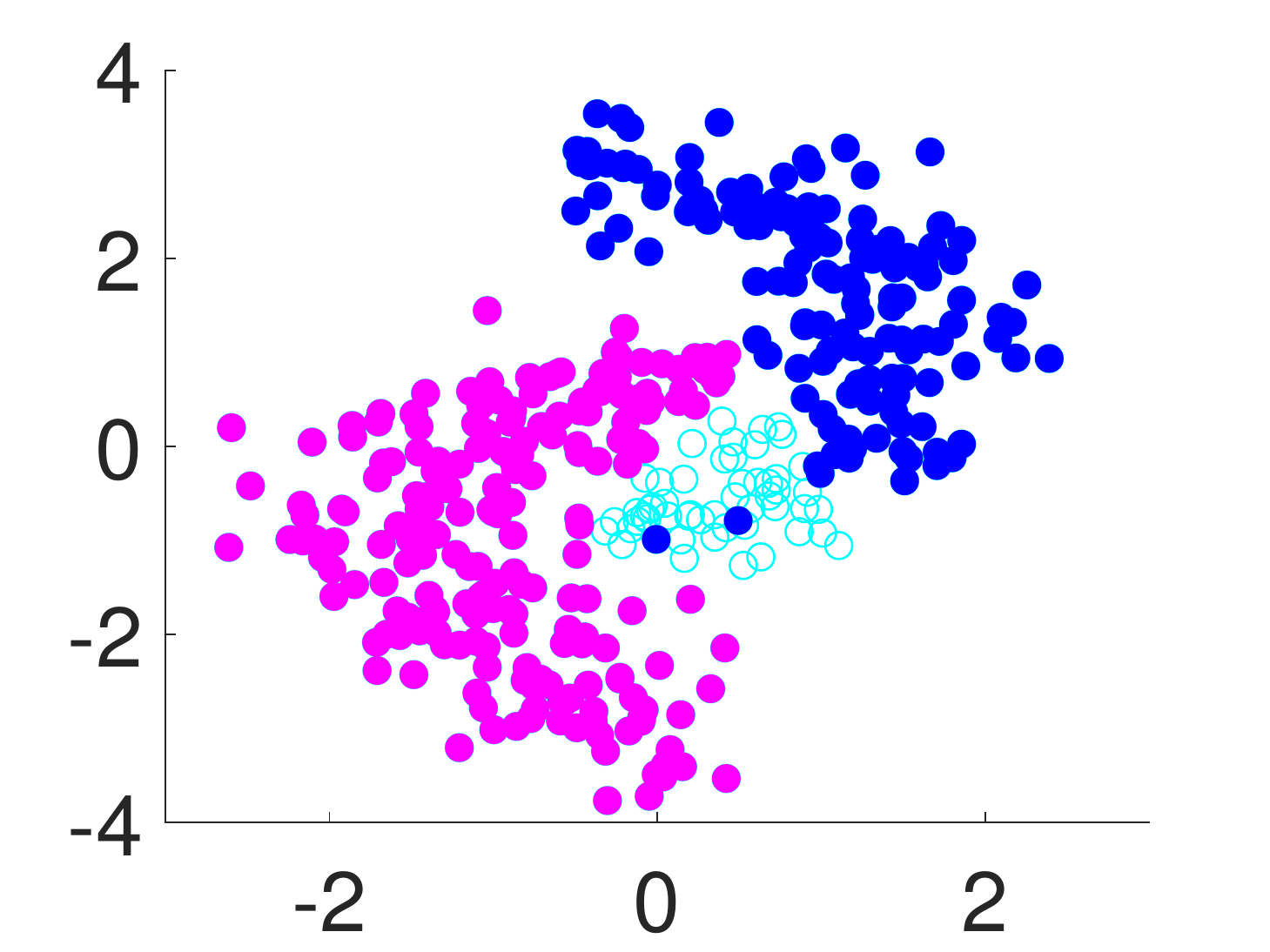}}
		\subfigure[after $312$-th iteration]{\includegraphics[scale=0.20]{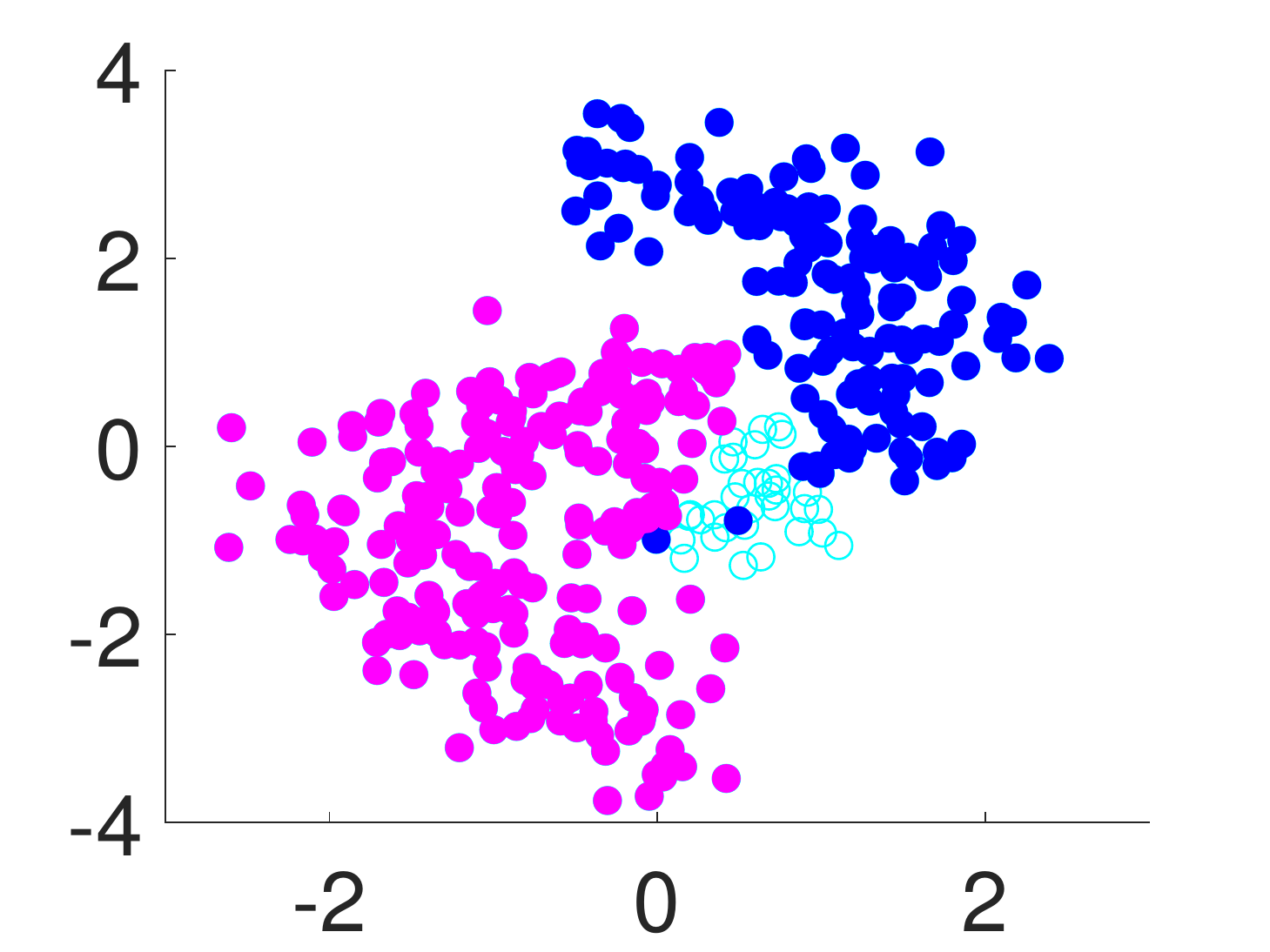}}
		\subfigure[after $333$-th iteration]{\includegraphics[scale=0.20]{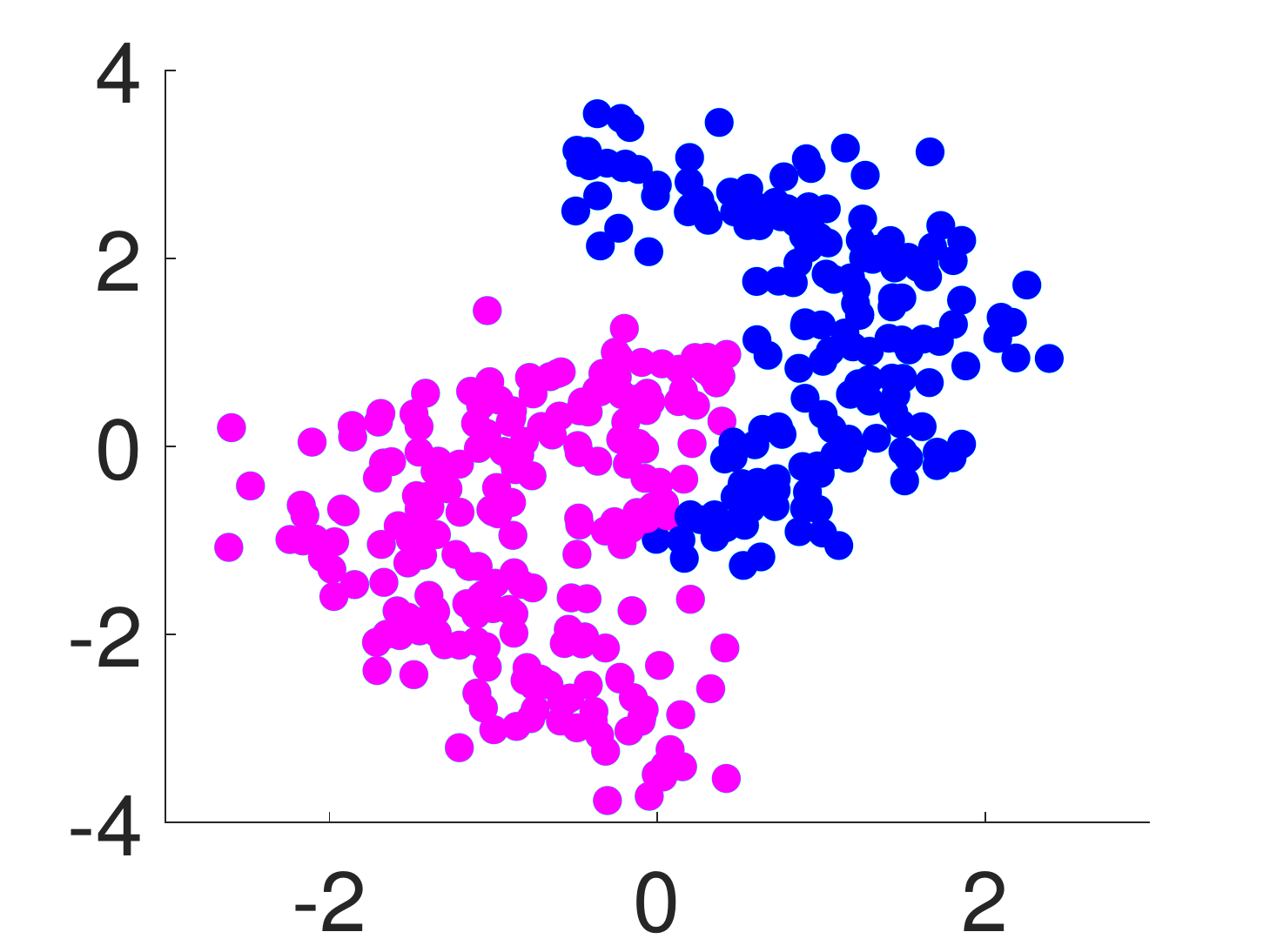}}
		\caption{The visualization of the screening process of IAES on two-moons with $p=400$. The filled dots in magenta and blue are the identified active and inactive elements, respectively. The unfilled dots in cyan are the unidentified elements.}
		\label{fig:screening-results-p-200}
	\end{center}
\end{figure*}

Figure \ref{fig:screening-results-p-200} visualizes the screening process of IAES on two-moons when $p=400$. It shows that, during the optimization process, IAES identifies the elements that are easy to be classified first and then identifies the rest.

Table \ref{table:run-time-syn} reports the running time of MinNorm without and with AES (\ref{rule:AES-v1} + \ref{rule:AES-v2}), IES (\ref{rule:IES-v1} + \ref{rule:IES-v2}) and IAES for solving the problem \ref{eqn:SFM} on two-moons. We can see that the speedup of IAES can be up to 10 times. In all the datasets, IAES is significantly faster than MinNorm, MinNorm with AES or IES. At last, we can see that the time costs of AES, IES and IAES are negligible.

\begin{table*}[htb!]
	\centering
	\caption{Running time (in seconds) for solving \ref{eqn:SFM} on two-moons.}
	\label{table:run-time-syn}
	{\footnotesize
		\begin{tabular}{|p{1.3cm}<{\centering}|p{1.2cm}<{\centering}|p{0.5cm}<{\centering}|p{1.2cm}<{\centering}|p{1.0cm}<{\centering}|p{0.5cm}<{\centering}|p{1.2cm}<{\centering}|p{1.0cm}<{\centering}|p{0.6cm}<{\centering}|p{1.2cm}<{\centering}|p{1.0cm}<{\centering}|}
			\hline
			\multirow{2}{*}{Data} & \multirow{2}{*}{MinNorm} & \multicolumn{3}{c|}{AES+MinNorm} & \multicolumn{3}{c|}{IES+MniNorm} & \multicolumn{3}{c|}{IAES+MinNorm} \\ \cline{3-11} 
			&  & AES & MinNorm & Speedup & IES & MinNorm & Speedup & IAES & MinNorm & Speedup \\ \hline
			$p=200$ &29.1&0.08&12.5&2.3 &0.07&12.2&2.4&0.10& 4.3&\textbf{6.8}  \\ \hline
			$p=400$ &829.1&0.09&106.5&2.8 &0.12&231.7&3.6 &0.15&82.7&\textbf{10.0}  \\ \hline
			$p=600$ &2,084.5&0.12&408.7&5.1&0.13&671.1&3.1 &0.18 & 217.1&\textbf{9.6}  \\ \hline
			$p=800$ &2701.1&0.15 &534.0 &5.1 & 0.13&998.9&2.7 &0.25& 400.2&\textbf{6.8} \\ \hline
			$p=1000$ &5422.9&0.20&1177.4&4.6 &0.19 &1453.5&3.7 &0.30 &774.7 &\textbf{7.0} \\ \hline
		\end{tabular}
	}
\end{table*}

\subsection{Experiments on Real Datasets}
In this experiment, we evaluate the performance of IAES on an image segmentation task.  We use five images (included in the supplemental material) in \cite{rother2004grabcut} to evaluate IAES. The objective function $F(A)$ is the sum of the unary potentials for all individual pixels and the pairwise potentials of a $8$-neighbor grid graph: 
\begin{align}
F(A) = \u(A) +\sum_{i\in A, j \in V/A} d(i,j), \nonumber 
\end{align}
where $V$ presents all the pixels, $\u \in \R^V$ is the unary potential derived from the Gaussian Mixture model \cite{rother2004grabcut}, and $d(i,j) = \exp \{ -\|\x_i-\x_j\|^2\}$ ($\x_i$ and $\x_j$ are the values of two pixels) if $i,j$ are neighbors, otherwise $d(i,j)=0$. Table \ref{data_description} provides the statistics of the resulting image segmentation problems, including the numbers of the pixels and the edges in the $8$-neighbor grid graph.

\begin{table}[htb!]
	\begin{minipage}{1\linewidth}
		\begin{center}
			\caption{Statistics of the image segmentation problems.}\label{data_description}
			\begin{footnotesize}
				\begin{tabular}{|l|l|l|l|}
					\hline
					image & \#pixels & \#edges\\ \hline		
					image1 &50,246&201,427 \\ \hline
					image2 &26,600&106,733 \\ \hline
					image3 &51,000&204,455 \\ \hline
					image4 &60,000&240,500 \\ \hline
					image5 &45,200&181,226 \\ \hline
				\end{tabular}
			\end{footnotesize}
		\end{center}
	\end{minipage}
\end{table}

\begin{figure*}[htb!]
	\begin{center}
		\subfigure[image1]{\includegraphics[scale=0.20]{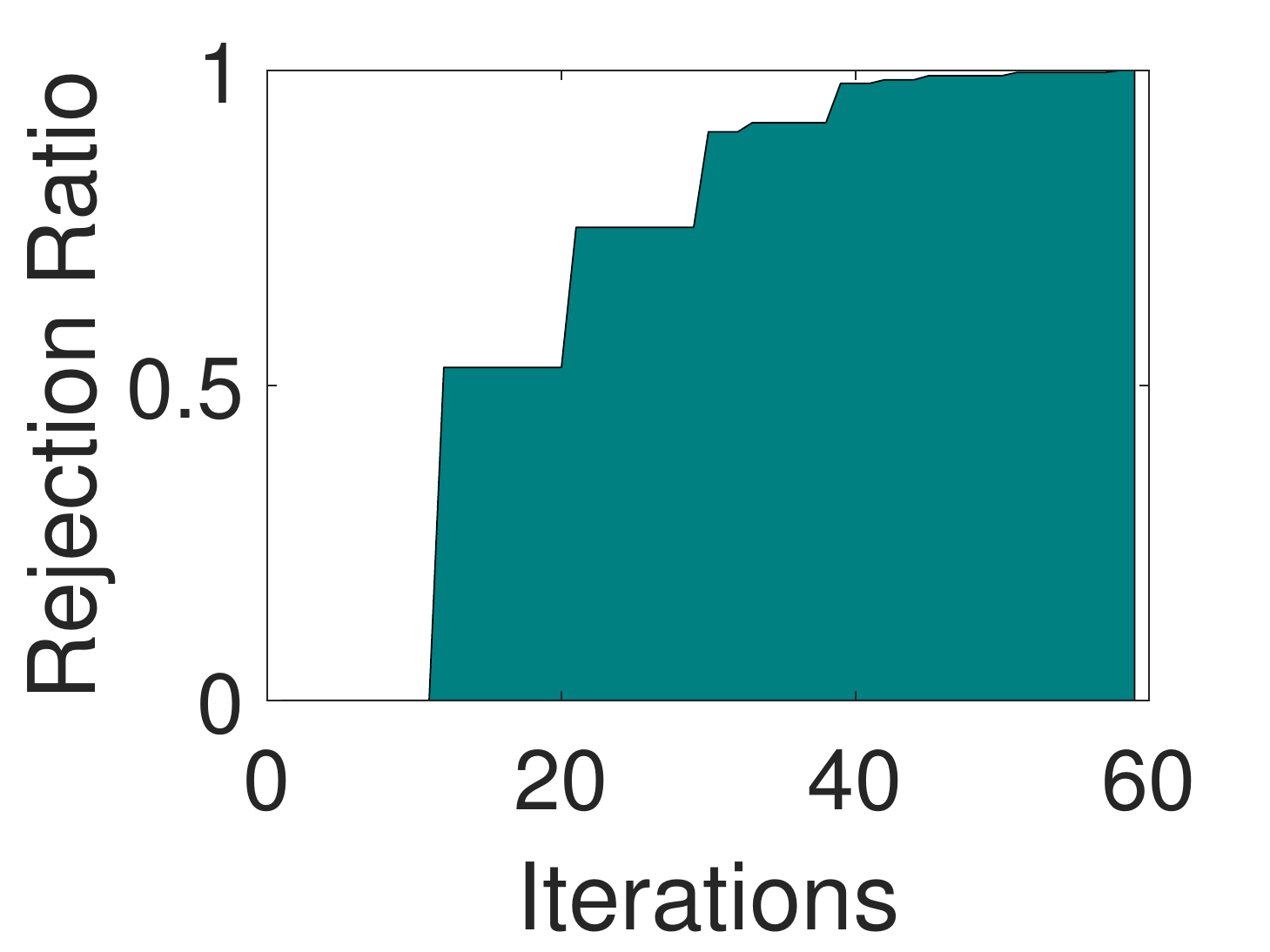}}
		\subfigure[image2]{\includegraphics[scale=0.20]{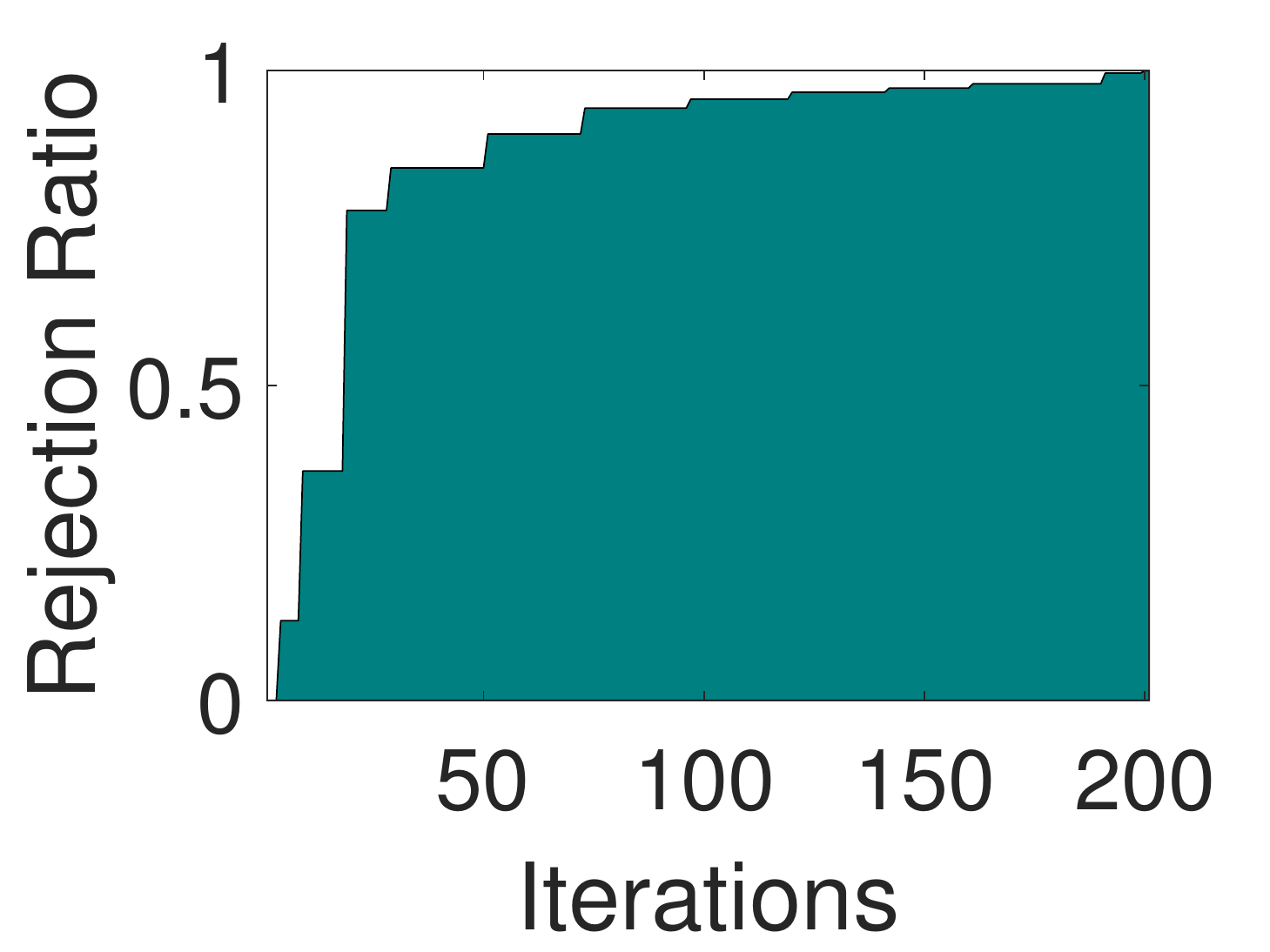}}
		\subfigure[image3]{\includegraphics[scale=0.20]{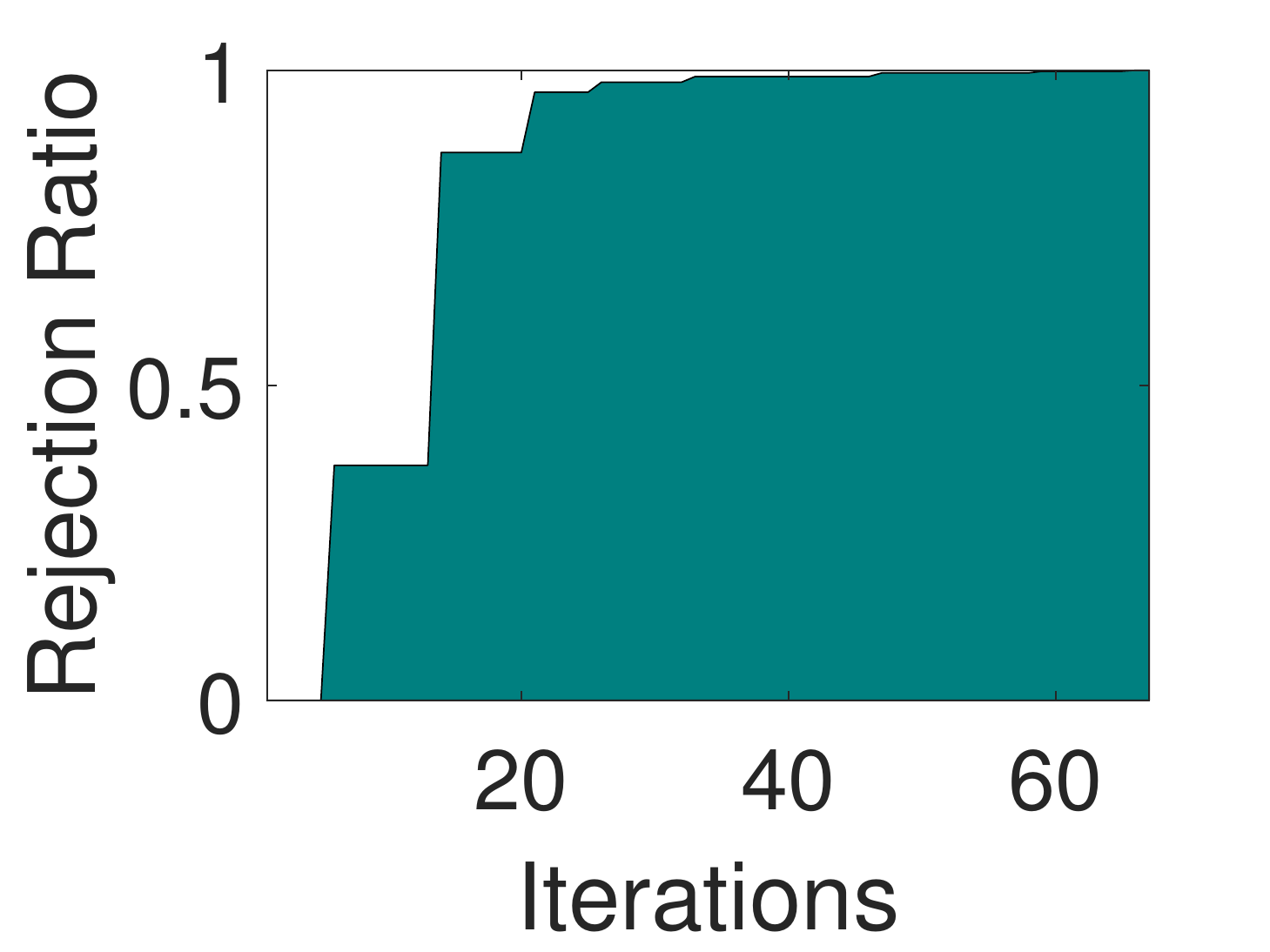}}
		\subfigure[image4]{\includegraphics[scale=0.20]{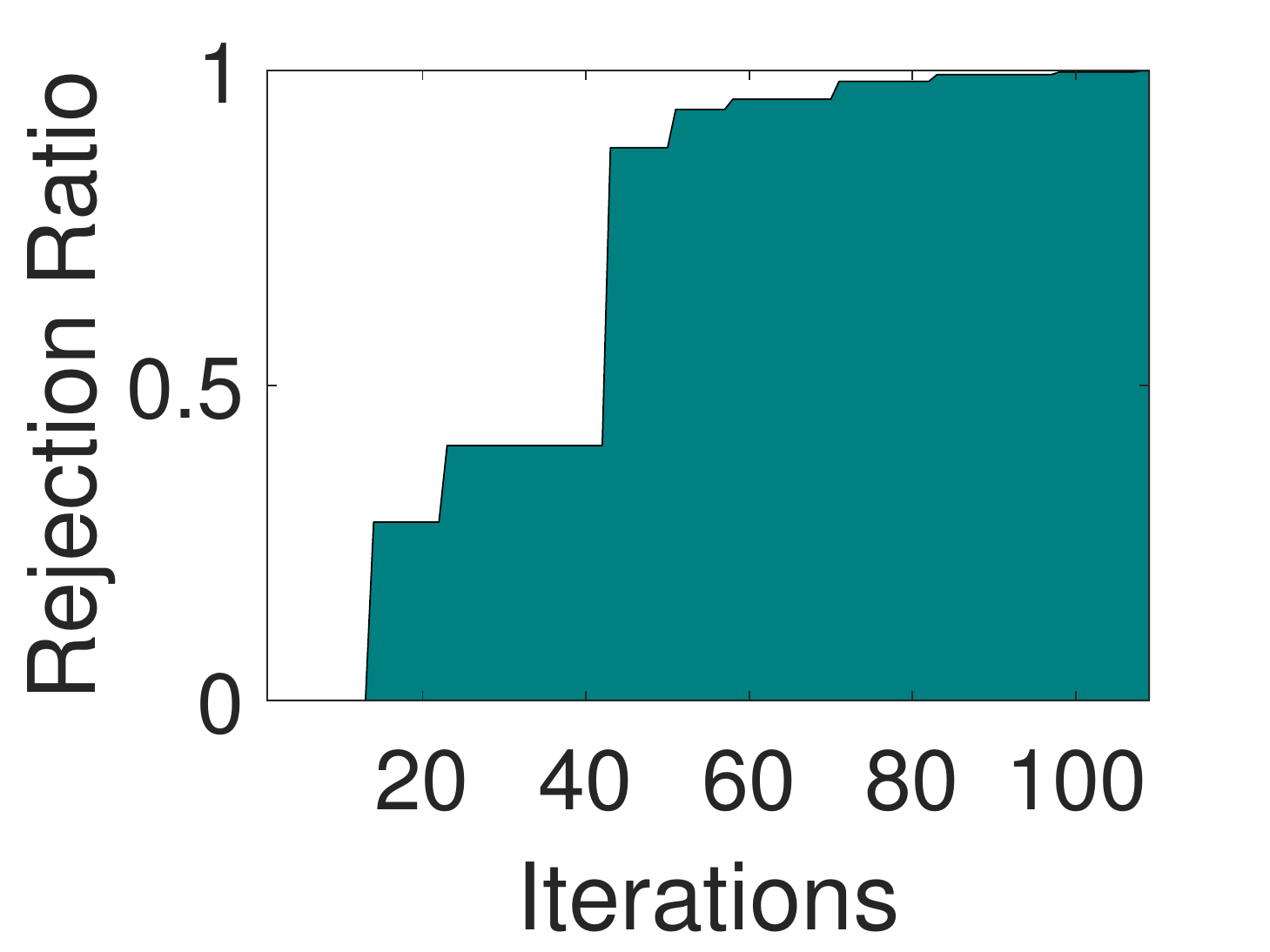}}
		\subfigure[image5]{\includegraphics[scale=0.20]{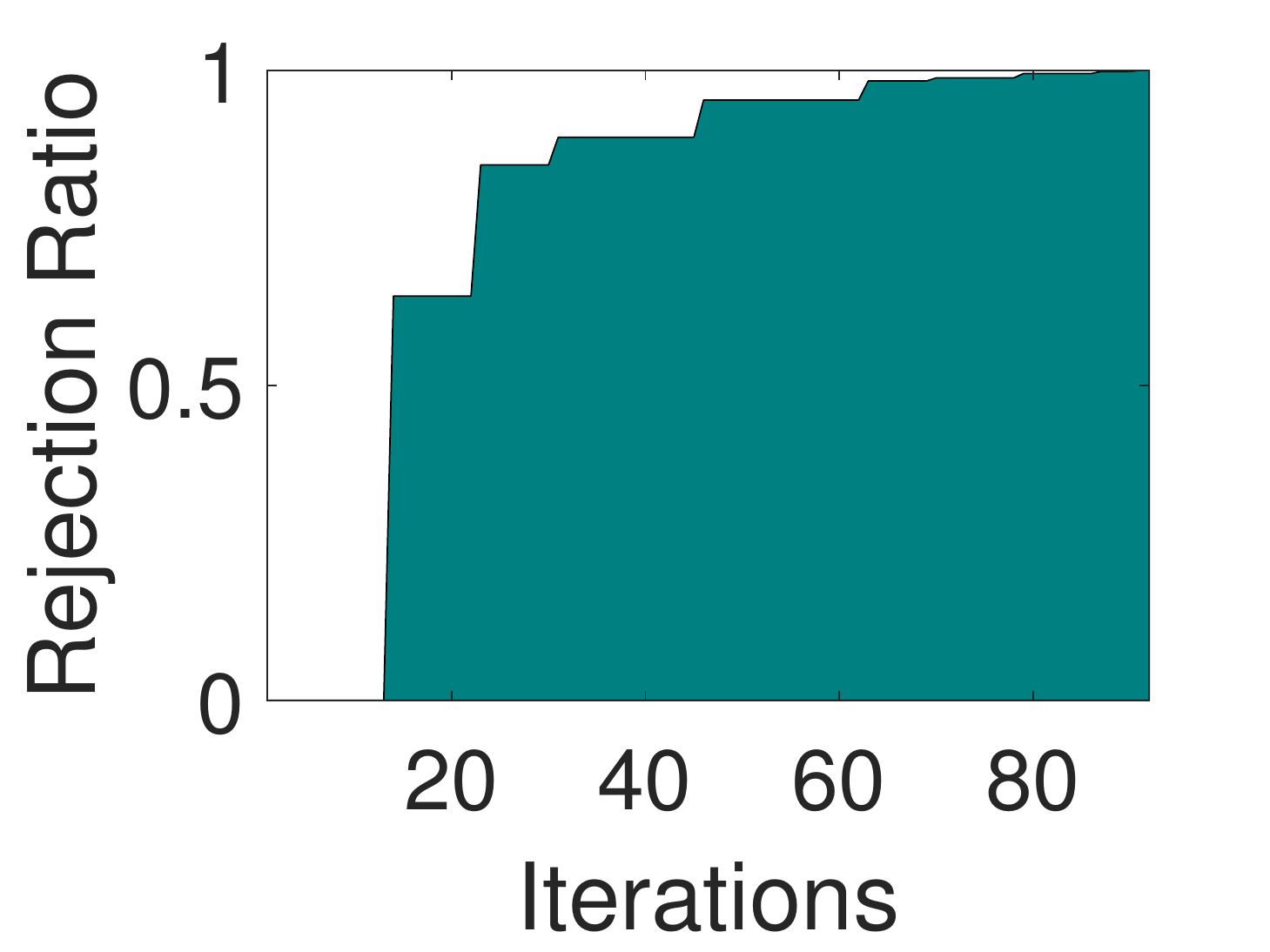}}
		\caption{Rejection ratios of IAES on real datasets over the iterations}
		\label{fig:reject-ratios-real}
	\end{center}
\end{figure*}

The rejection ratios in Figure \ref{fig:reject-ratios-real} show that IAES can identify the active and inactive elements during the optimization process incrementally until all of them are identified. This implies that IAES can lead to a significant speedup in the time cost. 

Table \ref{table:run-time-real} reports the detailed time cost of MinNorm without and with AES, IES and IAES for solving the image segmentation problems. We can see that IAES leads to significant speedups, which are up to 30.7 times. In addition, we notice that the speedup gained by AES is small. The reason is that AES is used to identify the pixels of the foreground, which is a small region in the image, and thus the problem size cannot be reduced dramatically even if all the active elements are identified.  
\begin{table*}[htb!]
	\centering
	\caption{Running time (in seconds) for solving \ref{eqn:SFM} on the task of image segmentation.}
	\label{table:run-time-real}
	{\footnotesize
		\begin{tabular}{|p{1.3cm}<{\centering}|p{1.2cm}<{\centering}|p{0.5cm}<{\centering}|p{1.2cm}<{\centering}|p{1.0cm}<{\centering}|p{0.5cm}<{\centering}|p{1.2cm}<{\centering}|p{1.0cm}<{\centering}|p{0.6cm}<{\centering}|p{1.2cm}<{\centering}|p{1.0cm}<{\centering}|}
			\hline
			\multirow{2}{*}{Data} & \multirow{2}{*}{MinNorm} & \multicolumn{3}{c|}{AES+MinNorm} & \multicolumn{3}{c|}{IES+MniNorm} & \multicolumn{3}{c|}{IAES+MinNorm} \\ \cline{3-11} 
			&  & AES & MinNorm & Speedup & IES & MinNorm & Speedup & IAES & MinNorm & Speedup \\ \hline
			image1 &1575.6&0.10&1412.8 &1.12&0.11&242.7&6.48&0.21&78.3& \textbf{20.19} \\ \hline
			image2 &1780.6&0.21&1201.6&1.48&0.30&616.6 &2.89 &0.50 &130.1&\textbf{13.69} \\ \hline
			image3 &6775.8&0.51&5470.8&1.24&0.53&1080.4&6.27&0.17&220.7&\textbf{30.70} \\ \hline
			image4 &6613.5&0.42&5773.1&1.15&0.43&1286.3&5.14&0.32&553.2&\textbf{11.96} \\ \hline
			image5 &4025.4&0.40&3638.8&1.11&0.17&506.4&7.95 &0.51&187.5& \textbf{21.50}\\ \hline
		\end{tabular}
	}
\end{table*}

At last, from Table \ref{table:run-time-real}, we can also see that the speedup we achieve is supper-additive (speedup of AES + speedup of IES $<$ speedup of IAES). This can usually be expected, which comes from the super linear computational complexity of each iteration in MinNorm, leading to a super-additive saving in the computational cost.  We notice that the speedup we achieve on some of the two-moon datasets is not super-additive. The reason is that we cannot identify a lot of elements in the early stage (Figure \ref{fig:reject-ratios-syn}). Thus, the early stage takes up too much time cost.

\section{Conclusion}\label{sec:conclusion}
In this paper, we proposed a novel safe element screening method IAES for SFM to accelerate its optimization process by simultaneously identifying the active and inactive elements. Our major contribution is a novel framework for accurately estimating the optimum of the corresponding primal problem of SFM developed by carefully studying the strong convexity of the primal and dual problems, the structure of the base polyhedra, and the optimality conditions of SFM. To the best of our knowledge, IAES is the first screening method in the fields of SFM and even combinatorial optimization. The extensive experimental results demonstrate that IAES can achieve significant speedups.

\appendix
%\vspace{0.5in}

%\begin{center}
%\line(1,0){250}
%\end{center}
%\begin{center}
%	\rule{14cm}{4.pt}\\
%	{\Large\bf Scaling Up Sparse Support Vector Machine\\by Simultaneous Feature and Sample Reduction\\ \vspace{2mm}Supplementary Material }
%	\rule{14cm}{0.7pt}
%\end{center}

%In this supplement, we present all of the details we mentioned in the main text.

\section{Appendix}
	In this supplement, we present the detailed proofs of all the theorems in the main text.
\subsection{Proof of Theorem \ref{thm:dual-kkt}}
\proof of Theorem \ref{thm:dual-kkt}:

(i) Since $f(\w) = \max_{\s\in B(F)} \langle \w, \s \rangle$, we can have
\begin{align}
&\min_{ \w \in \R^p} f(\w) + \sum_{j=1}^{p}\psi_j(\mbox{[}\w\mbox{]}_j) \label{eqn:thm-dual-kkt-proof-3}\\
=& \min_{ \w \in \R^p} \max_{\s\in B(F)} \langle \w, \s \rangle+ \sum_{j=1}^{p}\psi_j(\mbox{[}\w\mbox{]}_j) \nonumber  \\
=& \max_{\s\in B(F)} \min_{ \w \in \R^p} \langle \w, \s \rangle+ \sum_{j=1}^{p}\psi_j(\mbox{[}\w\mbox{]}_j) \label{eqn:thm-dual-kkt-proof-1}\\
=& \max_{\s\in B(F)} -\sum_{j=1}^{p}\psi^*_j(-\mbox{[}\s\mbox{]}_j),\label{eqn:thm-dual-kkt-proof-2}
\end{align}
where (\ref{eqn:thm-dual-kkt-proof-1}) holds due to the strong duality theorem \cite{borwein2010convex}, and (\ref{eqn:thm-dual-kkt-proof-2}) holds due to the definitions of the Fenchel conjugate of $\psi_j$.

(ii) From (\ref{eqn:thm-dual-kkt-proof-3}), we have
\begin{align}
&\s^* \in \arg \max_{\s\in B(F)} \langle \w^*, \s \rangle \nonumber \\
&\Leftrightarrow \langle \w^*, \s^* \rangle\geq  \langle \w^*, \s \rangle, \forall \s\in B(F) \nonumber \\
&\Leftrightarrow  \w^* \in N_{B(F)}(\s^*). \nonumber 
\end{align}
From Eq. (\ref{eqn:thm-dual-kkt-proof-2}), we have
\begin{align}
&\w^* \in \arg \min_{ \w \in \R^p} \langle \w, \s^* \rangle+ \sum_{j=1}^{p}\psi_j(\mbox{[}\w\mbox{]}_j)\nonumber \\
& \Leftrightarrow \mbox{[}\s\mbox{]}_k^* \in -\partial \psi_k(\mbox{[}\w\mbox{]}_k^*), \forall k \in V.\nonumber
\end{align}
The proof is complete. 
\endproof
\subsection{Proof of Lemma \ref{lemma:scaled-problem}}
\proof of Lemma \ref{lemma:scaled-problem}:

(i) It is the immediate conclusion of Theorem \ref{thm:convex-sfm}.

(ii) Since $\hcalE \subseteq A^*$ and $\hcalG \subseteq V / A^*$, we can solve the problem \ref{eqn:SFM} by fixing the set  $\hcalE$ and optimizing over $V/(\hcalE \cup \hcalG)$. And the objective function becomes $\hat{F}(C) := F(\hcalE \cup C) - F(\hcalE)$ with $C\subseteq V/(\hcalE \cup \hcalG)$. Thus, \ref{eqn:SFM} can be deduced to 
\begin{align}
\min_{C \subseteq V/(\hcalE\cup \hcalG)} \hat{F}(C) := F(\hcalE \cup C) - F(\hcalE). \nonumber
\end{align}

The second term of the new objective function $\hat{F}(C)$ is added to make $\hat{F}(\emptyset) = 0$, which is essential in submodular function analysis, such as Lov\'{a}sz extension, submodular and base polyhedra. 

Below, we argue that $\hat{F}(C)$ is a submodular function. 

For all $S \subseteq V/(\hcalE\cup \hcalG)$ and $T \subseteq V/(\hcalE\cup \hcalG)$, we have 
\begin{align}
\hat{F}(S) + \hat{F}(T) &= (F(\hcalE \cup S) - F(\hcalE)) + (F(\hcalE \cup T) - F(\hcalE))\nonumber \\
&= F(\hcalE \cup S)+ F(\hcalE \cup T) - 2F(\hcalE)\nonumber \\
&\geq F((\hcalE \cup S)\cup (\hcalE \cup T))+F((\hcalE \cup S)\cap (\hcalE \cup T))- 2F(\hcalE)\label{eqn:lemma-scaled-problem-1} \\
& = F(\hcalE \cup (S \cup T)) + F(\hcalE\cup (S\cup T))- 2F(\hcalE)\nonumber \\
& = (F(\hcalE \cup (S \cup T))-F(\hcalE)) + (F(\hcalE\cup (S\cup T))- F(\hcalE)) \nonumber\\
& = \hat{F}(S\cup T) + \hat{F}(S \cap T).\nonumber
\end{align}
The inequality (\ref{eqn:lemma-scaled-problem-1}) comes from the submoduality of $F$. 

(iii) It is the immediate conclusion of (ii).  

The proof is complete. 
\endproof
\subsection{Proof of Theorem \ref{thm:optimum-estimation-v1}}

To prove Theorem \ref{thm:optimum-estimation-v1}, we need the following Lemma. 
\begin{lemma}\label{lemma:dual-of-submodular}\textup{[Dual of minimization of submodular functions, Proposition 10.3 in \cite{bach2013learning}]} Let $F$ be a submodular function such that $F(\emptyset) = 0$. We have:
	\begin{align}
	\min_{A\subseteq V}F(A) = \max_{\s \in B(F)} \s_{-}(V) = \frac{1}{2}\Big(F(V)-\min_{ \s \in B(F)} \|\s\|_1\Big), \label{eqn:dual-SFM-lemma}
	\end{align}
	where $\textup{[}\s_{-}\textup{]}_k = \min \{\textup{[} \s\textup{]}_k, 0\}$ for $\forall k \in V$. 
\end{lemma}

We now turn to prove Theorem \ref{thm:optimum-estimation-v1}.
\proof of Theorem \ref{thm:optimum-estimation-v1}:

Since $\hat{P}(\hat{\w})$ is $1$-strongly convex, for any $\hat{\w}\in dom \hat{P}(\hat{\w})$ and $\hat{\w}^* = \arg \min_{ \hat{\w} \in \R^{\hat{p}}}\hat{P}(\hat{\w})$, we can have
\begin{align}
\hat{P}(\hat{\w}) \geq \hat{P}(\w^*) + \langle \hat{\g}, \hat{\w} - \hat{\w}^* \rangle + \frac{1}{2}\|\hat{\w} - \hat{\w}^* \|_2^2, \nonumber 
\end{align}
where $\g \in \partial \hat{P}(\hat{\w}^*)$.

Since $dom \hat{P}(\hat{\w}) = \R^{\hat{p}}$, it holds that $0 \in \partial \hat{P}(\hat{\w}^*)$. Hence, we can obtain
\begin{align}
\frac{1}{2}\|\hat{\w} - \hat{\w}^* \|_2^2 \leq \hat{P}(\hat{\w}) - \hat{P}(\hat{\w}^*) . \nonumber 
\end{align}
In addition, we notice that $\hat{P}(\w^*) \geq \hat{D}(\hat{\s})$ for all $\hat{\s} \in B(\hat{F})$. By substituting this inequality into the above inequality, we obtain that 
\begin{align}
\frac{1}{2}\|\hat{\w} - \hat{\w}^* \|_2^2 \leq \hat{P}(\hat{\w}) - \hat{P}(\hat{\w}^*) \leq \hat{P}(\hat{\w}) - \hat{D}(\hat{\s})=G(\hat{\w}, \hat{\s}). \nonumber 
\end{align}
Thus,
\begin{align}
\hat{\w}^* \in \calB := \Big\{\w: \|\w - \hat{\w} \| \leq \sqrt{2G(\hat{\w}, \hat{\s})} \Big \}. \label{eqn:calB-proof}
\end{align}

According to the equation (\ref{opt-conditions}) in Theorem \ref{thm:dual-kkt}, we have that $-\hat{\w}^*$ is the optimal solution of the problem \ref{eqn:Q-D'}. Therefore, $-\hat{\w}^* \in B(\hat{F})$. From the definition of $B(\hat{F})$, we have
\begin{align}
-\langle \hat{\w}^*, \mathbf{1} \rangle = -\hat{\w}^*(\hat{V}) = \hat{F}(\hat{V}).\nonumber 
\end{align}
Thus, 

\begin{align}
\hat{\w}^* \in \calP := \Big\{\w: \langle \w, \mathbf{1} \rangle = -\hat{F}(\hat{V}) \Big \}. \label{eqn:P-proof}
\end{align}

By section 7.3 of \cite{bach2013learning}), it holds that the unique minimizer of problem \ref{eqn:Q-D'} is also a maximizer of 
\begin{align}
\max_{\s \in B(\hat{F})}\s_{-}(V). \nonumber
\end{align} 
Hence, it holds that
\begin{align}
\| \hat{\s}^* \|_1 \leq \|\hat{\s}\|_1 \mbox{ for all } \hat{\s} \in B(\hat{F}).\label{eqn:thm:optimum-estimation-v1-1} 
\end{align}
From Lemma \ref{lemma:dual-of-submodular}, we have
\begin{align}
\hat{F}(C)\geq \frac{1}{2}(\hat{F}(\hat{V})-\| \hat{\s}\|_1), \mbox{for all } \hat{\s} \in B(\hat{F}), \nonumber \\
\Rightarrow  \| \hat{\s}\|_1 \geq \hat{F}(\hat{V})- 2\hat{F}(C) , \mbox{for all } \hat{\s} \in B(\hat{F}). \label{eqn:thm:optimum-estimation-v1-2}  
\end{align}
By combining (\ref{eqn:thm:optimum-estimation-v1-1}) and (\ref{eqn:thm:optimum-estimation-v1-2}), we acquire
\begin{align}
\hat{F}(\hat{V})- 2\hat{F}(C) \leq \|\hat{\s}^* \|_1 \leq \|\hat{\s} \|_1, \mbox{for all } \hat{\s} \in B(\hat{F}). \nonumber
\end{align}
Since $\hat{\w}^* = -\hat{\s}^*$, we have 
\begin{align}
\hat{F}(\hat{V})- 2\hat{F}(C) \leq \|\hat{\w}^* \|_1 \leq \|\hat{\s} \|_1, \mbox{for all } \hat{\s} \in B(\hat{F}). \nonumber
\end{align}
Thus, we obtain
\begin{align}
\hat{\w}^* \in \Omega := \Big\{\w: \hat{F}(\hat{V}) - 2\hat{F}(C) \leq \|\w \|_1 \leq \|\hat{\s}\|_1 \Big \}. \label{eqn:Omega-proof}
\end{align}
From (\ref{eqn:calB-proof}), (\ref{eqn:P-proof}) and (\ref{eqn:Omega-proof}), we have $\hat{\w}^* \in \calB \cap \Omega \cap \calP$. 

The proof is complete. 
\endproof

\subsection{Proof of Lemma \ref{lemma:upper-lower-v1}}
\proof of Lemma \ref{lemma:upper-lower-v1}: 

For any $j = 1, ..., \hat{p}$, we have
\begin{align}
&\sum_{i\neq j}(\mbox{[}\w\mbox{]}_i-\mbox{[}\hat{\w}\mbox{]}_i)^2 \leq 2G(\hat{\w},\hat{\s}) -(\mbox{[}\w\mbox{]}_j - (\mbox{[}\hat{\w}\mbox{]}_j)^2, \label{eqn:1-v1} \\
&\sum_{i\neq j} \mbox{[}\w\mbox{]}_i = -\hat{F}(\hat{V})-\mbox{[}\w\mbox{]}_j. \label{eqn:2-v1}
\end{align}

By fixing the component $\mbox{[}\w\mbox{]}_j$, we can see that (\ref{eqn:1-v1}) and (\ref{eqn:2-v1}) are a  ball and a plane in $\R^{\hat{p}-1}$, respectively. To make the intersection of (\ref{eqn:1-v1}) and (\ref{eqn:2-v1}) non-empty, we just need to restrict the distance between the center of ball (\ref{eqn:1-v1}) and the plane (\ref{eqn:2-v1}) smaller than the radius, {\it{i.e.}}, 
\begin{align}
\frac{|\sum_{i\neq j} \mbox{[}\hat{\w}\mbox{]}_i + \hat{F}(\hat{V})+\mbox{[}\w\mbox{]}_j|}{\sqrt{\hat{p}-1}} \leq \sqrt{2G(\hat{\w}, \hat{\s})-(\mbox{[}\w\mbox{]}_j - \mbox{[}\hat{\w}\mbox{]}_j)^2}, \nonumber 
\end{align}
which is equivalent to 
\begin{align}
&\hat{p} \mbox{[}\w\mbox{]}_j^2 + b\mbox{[}\w\mbox{]}_j + c \leq 0, \label{eqn:3-v1} \\
&\mbox{where } b =2\Big(\sum_{i\neq j} \mbox{[}\hat{\w}\mbox{]}_i+\hat{F}(\hat{V})-(\hat{p}-1)\mbox{[}\hat{\w}\mbox{]}_j\Big),\nonumber \\
&\mbox{ and } c =\Big(\sum_{i\neq j} \mbox{[}\hat{\w}\mbox{]}_i + \hat{F}(\hat{V})\Big)^2- 2(\hat{p}-1)G(\hat{\w}, \hat{\s})+(\hat{p}-1)\mbox{[}\hat{\w}\mbox{]}_j^2. \nonumber  
\end{align}
Thus we have

\begin{align}
\mbox{[}\w\mbox{]}_j \in \mathbf{[}\frac{-b - \sqrt{b^2 - 4\hat{p}c}}{2\hat{p}}, \frac{-b + \sqrt{b^2 - 4\hat{p}c}}{2\hat{p}} \mathbf{]} \nonumber 
\end{align}
At last, we would point out here that since $\hat{\w}^*$ must be in the intersection of the ball (\ref{eqn:1-v1}) and the plane (\ref{eqn:2-v1}). Hence, inequality (\ref{eqn:3-v1}) can be satisfied with $\mbox{[}\hat{\w}^*\mbox{]}_j$, which implies that  $b^2 - 4\hat{p}c$ would never be negative.

The proof is complete. 
\endproof

\subsection{Proof of Theorem \ref{thm:IAES-v1}}
\proof of Theorem \ref{thm:IAES-v1}:

(i): According to $\min_{\w\in \calB \cap \calP} \mbox{[} \w \mbox{]}_j = \mbox{[} \w \mbox{]}_j^{\min} > 0$ and $\hat{\w}^* \in \calB \cap \calP$, we have
\begin{align}
\mbox{[}\hat{\w}^*\mbox{]}_j > 0. \nonumber
\end{align}
From Theorem \ref{thm:convex-sfm}, it holds that $j\in \arg \min_{C\subseteq \hat{V}}\hat{F}(C) \subseteq A^*$. 

(ii): Since $\max_{\w\in \calB \cap \calP} \mbox{[} \w \mbox{]}_j = \mbox{[} \w \mbox{]}_j^{\max} < 0$ and $\hat{\w}^* \in \calB \cap \calP$, we have
\begin{align}
\mbox{[}\hat{\w}^*\mbox{]}_j < 0. \nonumber
\end{align}
From Theorem \ref{thm:convex-sfm}, we have $j \notin \arg \min_{C\subseteq \hat{V}} \hat{F}(C)$.
Note that $A^*= \calE \cup \arg \min \hat{F}(C)$ and $j \notin \calE$. Therefore $j \notin A^*$. 

(iii) It is the immediate conclusion from (i) and (ii).

The proof is complete. 
\endproof
\subsection{Proof of Lemma \ref{lemma:upper-lower-v2}}	
\proof of Lemma \ref{lemma:upper-lower-v2}:

(i) We just need to prove that 
\begin{align}
\begin{cases}
\mbox{[}\w \mbox{]}_j^{\min} > 0 \mbox{ if } \mbox{[} \hat{\w} \mbox{]}_j > \sqrt{2G(\hat{\w}, \hat{\s})},\\
\mbox{[}\w \mbox{]}_j^{\max} < 0 \mbox{ if } \mbox{[} \hat{\w} \mbox{]}_j < -\sqrt{2G(\hat{\w}, \hat{\s})}.
\end{cases} \nonumber 
\end{align}
We divide the proof into two parts.
First, when $\mbox{[} \hat{\w} \mbox{]}_j > \sqrt{2G(\hat{\w}, \hat{\s})}$, considering the definition of $\mbox{[}\w \mbox{]}_j^{\min}$, we have
\begin{align}
\mbox{[}\w \mbox{]}_j^{\min} = \min_{\w\in \calB\cap \calP}\mbox{[}\w\mbox{]}_j \geq \min_{\w\in \calB}\mbox{[}\w\mbox{]}_j = \mbox{[} \hat{\w} \mbox{]}_j-\sqrt{2G(\hat{\w}, \hat{\s})} > 0. \nonumber 
\end{align}

In this case, the element $j$ can be screened by rule \ref{rule:AES-v1}. 

On the other hand, when $\mbox{[} \hat{\w} \mbox{]}_j < -\sqrt{2G(\hat{\w}, \hat{\s})}$, from the definition of $\mbox{[}\w \mbox{]}_j^{\max}$, we have
\begin{align}
\mbox{[}\w \mbox{]}_j^{\max} = \max_{\w\in \calB\cap \calP}\mbox{[}\w\mbox{]}_j \leq \max_{\w\in \calB}\mbox{[}\w\mbox{]}_j = \mbox{[} \hat{\w} \mbox{]}_j+\sqrt{2G(\hat{\w}, \hat{\s})} < 0. \nonumber 
\end{align}
In this case, the element $j$ can be screened by rule \ref{rule:IES-v1}. 

(ii) We note that the point $\v$ with $\mbox{[} \v \mbox{]}_j = 0$ and $\mbox{[} \v \mbox{]}_k = \mbox{[} \hat{\w} \mbox{]}_k$ for all $k \neq j, k = 1,2,..,\hat{p}$  belongs to the ball $\calB$. Thus, we have 
\begin{align}
\min_{\w \in \calB,  \mbox{[}\w\mbox{]}_j \leq 0}\|\w\|_1 \leq \sum_{i\neq j} |\mbox{[}\hat{\v}\mbox{]}_i| = \|\hat{\w}\|_1 - \mbox{[} \hat{\w} \mbox{]}_j < \|\hat{\w}\|_1. \nonumber 
\end{align}

Now, we turn to calculate $\max_{\w \in \calB, \mbox{[}\w\mbox{]}_j \leq 0}\|\w\|_1$. 

We note that the range of $\mbox{[}\w\mbox{]}_j$  is $\mbox{[} -\sqrt{2G(\hat{\w}, \hat{\s})}+ \mbox{[}\hat{\w}\mbox{]}_j, \sqrt{2G(\hat{\w}, \hat{\s})}+ \mbox{[}\hat{\w}\mbox{]}_j \mbox{]}$ when $\w \in \calB$. Hence, the problem $\max_{\w \in \calB, \mbox{[}\w\mbox{]}_j \leq 0}\|\w\|_1$ can be decomposed into 

\begin{align}
\max_{-\sqrt{2G(\hat{\w}, \hat{\s})}+ \mbox{[}\hat{\w}\mbox{]}_j \leq \alpha \leq 0}\Big\{ \max_{\w \in \calB, \mbox{[}\w\mbox{]}_j = \alpha}\| \w \|_1\Big \}.\nonumber
\end{align}

We assume $\mbox{[}\w\mbox{]}_j = \alpha$ with $-\sqrt{2G(\hat{\w}, \hat{\s})} + \mbox{[}\hat{\w}\mbox{]}_j \leq \alpha \leq 0$ and first consider the following problem,
\begin{align}
\max_{\w \in \calB, \mbox{[}\w\mbox{]}_j = \alpha}\| \w \|_1, \nonumber
\end{align}
which can be rewritten as
\begin{align}
&\max_{\mbox{[}\w\mbox{]}_i, i\neq j}  -\alpha + \sum_{i\neq j} |\mbox{[}\w\mbox{]}_i|\nonumber \\
&s.t. \sum_{i \leq \hat{p}, i\neq j} (\mbox{[}\w\mbox{]}_i - \mbox{[}\hat{\w}\mbox{]}_j)^2 \leq 2G(\hat{\w}, \hat{\s}) - (\alpha - \mbox{[}\hat{\w}\mbox{]}_j)^2.  \nonumber
\end{align}
It is easy to check that the optimal solution of the problem above is 
\begin{align}
\mbox{[}\w\mbox{]}_i = \mbox{[}\hat{\w}\mbox{]}_i + \mathbf{sign}(\mbox{[}\hat{\w}\mbox{]}_i)\sqrt{\frac{2G(\hat{\w}, \hat{\s}) - (\alpha - \mbox{[}\hat{\w}\mbox{]}_j)^2}{\hat{p}-1}}. \nonumber
\end{align}
The function $\mathbf{sign}(\cdot): \R \rightarrow 
\{-1,1\}$ above takes 1 if the argument is positive, otherwise takes -1.  
And the corresponding optimal value is 
\begin{align}
\max_{\w \in \calB, \mbox{[}\w\mbox{]}_j = \alpha}\| \w \|_1 = -\alpha + \sum_{i\neq j} |\mbox{[}\hat{\w}\mbox{]}_i| + \sqrt{\hat{p}-1} \sqrt{2G(\hat{\w}, \hat{\s}) - (\alpha - \mbox{[}\hat{\w}\mbox{]}_j)^2}. \nonumber  
\end{align}

Now, we denote $h(\alpha)= -\alpha + \sum_{i\neq j} |\mbox{[}\hat{\w}\mbox{]}_i| + \sqrt{\hat{p}-1} \sqrt{2G(\hat{\w}, \hat{\s}) - (\alpha - \mbox{[}\hat{\w}\mbox{]}_j)^2}$ and turn to solve 	

\begin{align}
\max_{-\sqrt{2G(\hat{\w}, \hat{\s})}+ \mbox{[}\hat{\w}\mbox{]}_j\leq \alpha \leq 0} h(\alpha) \nonumber
\end{align}
If $\mbox{[}\hat{\w}\mbox{]}_j - \sqrt{\frac{2G(\hat{\w}, \hat{\s})}{\hat{p}}} < 0 $, then 
\begin{align}
\max_{-\sqrt{2G(\hat{\w}, \hat{\s})}+ \mbox{[}\hat{\w}\mbox{]}_j\leq \alpha \leq 0} h(\alpha) &= h(\mbox{[}\hat{\w}\mbox{]}_j - \sqrt{\frac{2G(\hat{\w}, \hat{\s})}{\hat{p}}}) = -\mbox{[}\hat{\w}\mbox{]}_j + \sum_{i\neq j} |\mbox{[}\hat{\w}\mbox{]}_i| +\sqrt{2\hat{p}G(\hat{\w}, \hat{\s})}\nonumber \\
&= \|\hat{\w}\|_1-2\mbox{[}\hat{\w}\mbox{]}_j +\sqrt{2\hat{p}G(\hat{\w}, \hat{\s})}; \nonumber
\end{align}

else if $\mbox{[}\hat{\w}\mbox{]}_j - \sqrt{\frac{2G(\hat{\w}, \hat{\s})}{\hat{p}}} \geq 0$, then 
\begin{align}
\max_{-\sqrt{2G(\hat{\w}, \hat{\s})}+ \mbox{[}\hat{\w}\mbox{]}_j\leq \alpha \leq 0} h(\alpha) &= h(0) = \sum_{i\neq j} |\mbox{[}\hat{\w}\mbox{]}_i| +   \sqrt{\hat{p}-1} \sqrt{2G(\hat{\w}, \hat{\s}) -  \mbox{[}\hat{\w}\mbox{]}_j^2} \nonumber \\
&= \|\hat{\w}\|_1- \mbox{[}\w\mbox{]}_j +   \sqrt{\hat{p}-1} \sqrt{2G(\hat{\w}, \hat{\s}) -  \mbox{[}\hat{\w}\mbox{]}_j^2}.\nonumber
\end{align}
In a consequence, we have 	

\begin{align}
\max_{\w\in \calB, \mbox{[}\w\mbox{]}_j \leq 0} \|\w\|_1=  \begin{cases}
\|\hat{\w}\|_1-2\mbox{[}\hat{\w}\mbox{]}_j +\sqrt{2\hat{p}G(\hat{\w}, \hat{\s})}, &\mbox{if }  \mbox{[}\hat{\w}\mbox{]}_j - \sqrt{\frac{2G(\hat{\w}, \hat{\s})}{\hat{p}}} < 0\\
\|\hat{\w}\|_1- \mbox{[}\w\mbox{]}_j +   \sqrt{\hat{p}-1} \sqrt{2G(\hat{\w}, \hat{\s}) -  \mbox{[}\hat{\w}\mbox{]}_j^2}, &\mbox{ otherwise}.
\end{cases} \nonumber
\end{align}

(iii) Recall that the point $\v$ with $\mbox{[} \v \mbox{]}_j = 0$ and $\mbox{[} \v \mbox{]}_k = \mbox{[} \hat{\w} \mbox{]}_k$ for all $k \neq j, k = 1,2,..,\hat{p}$  lies in the ball $\calB$. Thus, we have 

\begin{align}
\min_{\w \in \calB,  \mbox{[}\w\mbox{]}_j \geq 0}\|\w\|_1 \leq \sum_{i\neq j} |\mbox{[}\hat{\v}\mbox{]}_i| = \|\hat{\w}\|_1 - \mbox{[} \hat{\w} \mbox{]}_j < \|\hat{\w}\|_1. \nonumber 
\end{align}

Now, we turn to calculate $\max_{\w \in \calB, \mbox{[}\w\mbox{]}_j \geq 0}\|\w\|_1$. 

We note that the range of $\mbox{[}\w\mbox{]}_j$ is $\mbox{[} -\sqrt{2G(\hat{\w}, \hat{\s})}+ \mbox{[}\hat{\w}\mbox{]}_j, \sqrt{2G(\hat{\w}, \hat{\s})}+ \mbox{[}\hat{\w}\mbox{]}_j \mbox{]}$  when $\w \in \calB$. Hence, we decompose the problem $\max_{\w \in \calB, \mbox{[}\w\mbox{]}_j \geq 0}\|\w\|_1$ into 
\begin{align}
\max_{0 \leq \alpha \leq \sqrt{2G(\hat{\w}, \hat{\s})}+ \mbox{[}\hat{\w}\mbox{]}_j}\Big\{ \max_{\w \in \calB, \mbox{[}\w\mbox{]}_j = \alpha}\| \w \|_1\Big \}.\nonumber 
\end{align}

We assume $\mbox{[}\hat{\w}\mbox{]}_j = \alpha$ with $0 \leq \alpha \leq \sqrt{2G(\hat{\w}, \hat{\s})}+ \mbox{[}\hat{\w}\mbox{]}_j$ and first solve the following problem:
\begin{align}
\max_{\w \in \calB, \mbox{[}\w\mbox{]}_j = \alpha}\| \w \|_1, \nonumber
\end{align}
which can be rewritten as
\begin{align}
&\max_{\mbox{[}\w\mbox{]}_i, i\neq j} \alpha + \sum_{i\neq j} |\mbox{[}\w\mbox{]}_i|\nonumber \\
&s.t. \sum_{i \leq \hat{p}, i\neq j} (\mbox{[}\w\mbox{]}_i - \mbox{[}\hat{\w}\mbox{]}_j)^2 \leq 2G(\hat{\w}, \hat{\s}) - (\alpha - \mbox{[}\hat{\w}\mbox{]}_j)^2.  \nonumber
\end{align}
It can be verified that the optimal solution of the problem above is 
\begin{align}
\mbox{[}\w\mbox{]}_i = \mbox{[}\hat{\w}\mbox{]}_i + \mathbf{sign}(\mbox{[}\hat{\w}\mbox{]}_i)\sqrt{\frac{2G(\hat{\w}, \hat{\s}) - (\alpha - \mbox{[}\hat{\w}\mbox{]}_j)^2}{\hat{p}-1}}. \nonumber
\end{align}
The function $\mathbf{sign}(\cdot): \R \rightarrow 
\{-1,1\}$ above takes 1 if the argument is positive, otherwise takes -1. And the corresponding optimal value is 
\begin{align}
\max_{\w \in \calB, \mbox{[}\w\mbox{]}_j = \alpha}\| \w \|_1 = \alpha + \sum_{i\neq j} |\mbox{[}\hat{\w}\mbox{]}_i| + \sqrt{\hat{p}-1} \sqrt{2G(\hat{\w}, \hat{\s}) - (\alpha - \mbox{[}\hat{\w}\mbox{]}_j)^2}. \nonumber  
\end{align}

Now, we denote $h(\alpha)= \alpha + \sum_{i\neq j} |\mbox{[}\hat{\w}\mbox{]}_i| + \sqrt{\hat{p}-1} \sqrt{2G(\hat{\w}, \hat{\s}) - (\alpha - \mbox{[}\hat{\w}\mbox{]}_j)^2}$ and turn to solve

\begin{align}
\max_{0 \leq \alpha \leq \sqrt{2G(\hat{\w}, \hat{\s})}+ \mbox{[}\hat{\w}\mbox{]}_j} h(\alpha) \nonumber
\end{align}
If $\mbox{[}\hat{\w}\mbox{]}_j + \sqrt{\frac{2G(\hat{\w}, \hat{\s})}{\hat{p}}} > 0 $, then 
\begin{align}
\max_{0 \leq \alpha \leq \sqrt{2G(\hat{\w}, \hat{\s})}+ \mbox{[}\hat{\w}\mbox{]}_j} h(\alpha) &= h(\mbox{[}\hat{\w}\mbox{]}_j + \sqrt{\frac{2G(\hat{\w}, \hat{\s})}{\hat{p}}}) = \mbox{[}\hat{\w}\mbox{]}_j + \sum_{i\neq j} |\mbox{[}\hat{\w}\mbox{]}_i| +\sqrt{2\hat{p}G(\hat{\w}, \hat{\s})}\nonumber \\
&= \|\hat{\w}\|_1+2\mbox{[}\hat{\w}\mbox{]}_j +\sqrt{2\hat{p}G(\hat{\w}, \hat{\s})}. \nonumber
\end{align}
Else if $\mbox{[}\hat{\w}\mbox{]}_j + \sqrt{\frac{2G(\hat{\w}, \hat{\s})}{\hat{p}}} \leq 0$, then 
\begin{align}
\max_{0 \leq \alpha \leq \sqrt{2G(\hat{\w}, \hat{\s})}+ \mbox{[}\hat{\w}\mbox{]}_j} h(\alpha) &= h(0) = \sum_{i\neq j} |\mbox{[}\hat{\w}\mbox{]}_i| +   \sqrt{\hat{p}-1} \sqrt{2G(\hat{\w}, \hat{\s}) -  \mbox{[}\hat{\w}\mbox{]}_j^2} \nonumber \\
&= \|\hat{\w}\|_1+ \mbox{[}\w\mbox{]}_j +   \sqrt{\hat{p}-1} \sqrt{2G(\hat{\w}, \hat{\s}) -  \mbox{[}\hat{\w}\mbox{]}_j^2}.\nonumber
\end{align}
Consequently, we have

\begin{align}
\max_{\w\in \calB, \mbox{[}\w\mbox{]}_j \geq 0} \|\w\|_1=  \begin{cases}
\|\hat{\w}\|_1+2\mbox{[}\hat{\w}\mbox{]}_j +\sqrt{2\hat{p}G(\hat{\w}, \hat{\s})}, &\mbox{if } \mbox{[}\hat{\w}\mbox{]}_j + \sqrt{\frac{2G(\hat{\w}, \hat{\s})}{\hat{p}}} > 0,\\
\|\hat{\w}\|_1+ \mbox{[}\w\mbox{]}_j +   \sqrt{\hat{p}-1} \sqrt{2G(\hat{\w}, \hat{\s}) -  \mbox{[}\hat{\w}\mbox{]}_j^2}, &\mbox{ otherwise}.
\end{cases} \nonumber
\end{align} 	
The proof is complete. 
\endproof
\subsection{Proof of Theorem \ref{thm:IAES-v2}}
\proof of Theorem \ref{thm:IAES-v2}: 

(i): Noting that
\begin{align}
\begin{cases}
0 < \mbox{[}\hat{\w}\mbox{]}_j \leq  \sqrt{2G(\hat{\w}, \hat{\s})}, \nonumber \\
\max_{\w\in \calB, \mbox{[}\w\mbox{]}_j \leq 0} \|\w\|_1 < \hat{F}(\hat{V})-2\hat{F}(C),  \nonumber
\end{cases}
\end{align}
and $\Omega = \Big\{\w: \hat{F}(\hat{V}) - 2\hat{F}(C) \leq \|\w \|_1 \leq \|\hat{\s}\|_1 \Big \}$, we have 
\begin{align}
\Big \{\w, \w\in \calB,  \mbox{[}\w\mbox{]}_j \leq 0 \Big\} \cap \Omega = \emptyset. \label{eqn:thm-IAES-v2-proof-1}
\end{align}
Since $\hat{\w}^* \in \calB \cap \Omega$, from (\ref{eqn:thm-IAES-v2-proof-1}) we have $\mbox{[} \hat{\w}^* \mbox{]}_j > 0$. 
Thus, from Theorem \ref{thm:convex-sfm} we have $j \in \arg \min \hat{F}(C)\subseteq A^*$. 

(ii): Since 
\begin{align}
\begin{cases}
-\sqrt{2G(\hat{\w}, \hat{\s})} \leq \mbox{[}\hat{\w}\mbox{]}_j < 0, \nonumber \\
\max_{\w\in \calB, \mbox{[}\w\mbox{]}_j \geq 0} \|\w\|_1 < \hat{F}(\hat{V})-2\hat{F}(C), \nonumber
\end{cases}
\end{align}
and  $\Omega = \Big\{\w: \hat{F}(\hat{V}) - 2\hat{F}(C) \leq \|\w \|_1 \leq \|\hat{\s}\|_1 \Big \}$, we have 
\begin{align}
\Big \{\w, \w\in \calB,  \mbox{[}\w\mbox{]}_j \geq 0 \Big\} \cap \Omega = \emptyset. \label{eqn:thm-IAES-v2-proof-2}
\end{align}
Since $\hat{\w}^* \in \calB \cap \Omega$, from (\ref{eqn:thm-IAES-v2-proof-2}) we have $\mbox{[} \hat{\w}^* \mbox{]}_j < 0$. 

From Theorem \ref{thm:convex-sfm}, we have $j \notin \arg \min_{C\subseteq \hat{V}} \hat{F}(C)$.
Noting that $A^*= \calE \cup \arg \min_{C\subseteq \hat{V}} \hat{F}(C)$ and $j \notin \calE$. Therefore $j \notin A^*$. 

(iii) It is the immediate conclusion of (i) and (ii). 

The proof is complete. 
\endproof

\bibliographystyle{plain}	
\bibliography{egbib}		
\end{document}